# Physics of Language Models: Part 2.1, Grade-School Math and the Hidden Reasoning Process

Tian Ye
tye2@andrew.cmu.edu
CMU / Meta FAIR

Zicheng Xu
zichengxu@meta.com
Meta FAIR

Yuanzhi Li
Yuanzhi.Li@mbzuai.ac.ae
MBZUAI

Zeyuan Allen-Zhu
zeyuanallenzhu@meta.com
Meta FAIR

July 31, 2024*

**Abstract**

Recent advances in language models have demonstrated their capability to solve mathematical reasoning problems, achieving near-perfect accuracy on grade-school level math benchmarks like GSM8K. In this paper, we formally study how language models solve these problems. We design a series of controlled experiments to address several fundamental questions: (1) Can language models truly develop reasoning skills, or do they simply memorize templates? (2) What is the model's hidden (mental) reasoning process? (3) Do models solve math questions using skills similar to or different from humans? (4) Do models trained on GSM8K-like datasets develop reasoning skills beyond those necessary for solving GSM8K problems? (5) What mental process causes models to make reasoning mistakes? (6) How large or deep must a model be to effectively solve GSM8K-level math questions?

Our study uncovers many hidden mechanisms by which language models solve mathematical questions, providing insights that extend beyond current understandings of LLMs.

---
*Project page + video: https://physics.allen-zhu.com/part-2-grade-school-math/part-2-1.
We would like to thank Lin Xiao, Chunting Zhou for many helpful conversations. We would like to extend special thanks to Lucca Bertoncini, Liao Hu, Caleb Ho, Wil Johnson, Apostolos Kokolis, Parth Malani, Alexander Miller, Junjie Qian and Shubho Sengupta from Meta FAIR; Henry Estela, Rizwan Hashmi, Lucas Noah, and Maxwell Taylor from the Meta Cloud Foundation team; as well as Ian Clark, Gourab De, Anmol Mann, and Max Pfeifer from W&B. Without their invaluable support, the experiments in this paper would not have been possible.

# 1 Introduction

The field of language models has made significant progress in recent years. Large models like GPT-4 [17] have shown initial signs of general intelligence [8], while smaller models have demonstrated good reasoning abilities by solving challenging coding and math problems [11, 15, 16].

In this paper, we focus on the ability of small language models to solve grade-school math problems. Unlike previous works that empirically push the accuracy of models on grade-school math benchmarks like GSM8K [9] and its augmentations (e.g., [16, 22]), we take a more principled approach. We aim to understand the following fundamental questions:

1. How do language models learn to solve grade-school level math problems? Do they just memorize templates, or do they learn reasoning skills similar to humans? Or do they discover new skills to solve the problems?

2. Do models trained *solely* on grade-school math problems only learn to solve these problems, or do they develop some more general intelligence?

3. How small can a language model be while still solving grade-school math problems? Is depth (number of layers) more important than width (number of neurons per layer), or does only size matter as suggested by practitioners [14]?

These questions are fundamental to understanding the intelligence of language models. To study them, it might seem tempting to start with a pre-trained model and fine-tune it on existing datasets like GSM8K or GPT-4 augmented ones (e.g., [16, 22]). However, this approach has significant limitations:

- DATA CONTAMINATION. The pretrain data of existing models mostly come from publicly available internet [10], which is a pile of mess. We do not know how many math problems are included or their structures. There is *significant concern regarding* whether the GSM8K benchmark has been *leaked to language models' training datasets* [22]. Even if the exact data is not, the pre-trained model might have seen almost identical questions (e.g., the same problem with different numbers). Thus, this approach cannot answer questions 1-3. We do not know whether a model truly learns the reasoning skills or it simply memorizes problem templates during training. Therefore, we **need full control over the model's pretrain data** and must train a language model from scratch. This point has been reiterated recently in [2, 3].

- SOLUTION DIVERSITY. The existing fine-tuning data, such as the GSM8K training set, contains only 7.5K grade-school math problems, which is insufficient to train a model from scratch. Although recent works use GPT-4 to augment GSM8K, this is not enough for our purpose. GPT-4 augmented problems might be biased towards a small number of solution templates, since the original GSM8K data has very few (obviously, at most 8K) solution templates. **We need a much larger, more diverse set of grade-school math problems**.

With these points in mind, we introduce our framework to generate a large set of diverse grade-school math (GSM) problems and use the dataset to train (from scratch) and test a GPT2-like language model. In the framework, we focus on the "logical reasoning" aspect of grade-school math problems, which involves the dependency of parameters in the problem statement, such as "Alice's apple is three times the sum of Bob's orange and Charles's banana." We use synthetic sentences to reduce the difficulty arising from *Common Sense*, like "a candle burned for 12 hours at 1 inch per hour" (implying the candle is reducing in length). We also remove the difficulty from pure arithmetic: we only consider integers and arithmetic mod23.[1]

Moreover, our framework ensures that the generated math problems are highly diverse and do not come from a small subset of templates. Even ignoring all the arithmetic, English, variable names, and unused parameters, our problems still have more than 90 trillion solution templates (see Proposition 2.2), much larger than the size of GPT2-small (100M). Thus, language models **cannot** solve the math problems in our case **by simply memorizing** the solution templates.

In this paper, we use the GPT2 model [18], but replace its positional embedding with rotary embedding (RoPE) [7, 20]. We still call it GPT2 for brevity. We summarize our main contributions:

- RESULT 2. We demonstrate that the GPT2 model, pretrained on our synthetic dataset, not only achieves 99% accuracy in solving math problems from the same distribution but also out-of-distribution generalizes, such as to those of longer reasoning lengths than any seen during training. This is similar to length generalization in arithmetics [6, 13], however, in our case, the model **has never seen *any* training example of the same length as in test time.** This signifies that the model can truly learn some reasoning skill instead of memorizing solution templates.

- RESULT 3. Crucially, the model can learn to generate shortest solutions, almost always avoiding *unnecessary* computations. This suggests that the model *formulates a plan* before it generates, in order to avoid computing any quantities that are not needed towards solving the underlying math problem.

- RESULT 4. We examine the model's internal states through probing, introducing six probing tasks to elucidate *how* the model solves math problems. For instance, we discover the model (mentally!) preprocesses the full set of necessary parameters before it starts any generation. Likewise, humans also do this preprocess although we write this down on scratch pads.

- RESULT 5. Surprisingly, the model also learns *unnecessary, yet important* skills after pretraining, such as all-pair dependency. Before any question is asked, it already (mentally!) computes with good accuracy which parameters depend on which, even though *some are not needed for solving the math problem*. Note that computing all-pair dependency is a skill **not needed** to fit all the solutions in the training data. To the best of our knowledge, this is the first evidence that a language model can **learn useful skills beyond** those necessary to fit its pretraining data.[2] This may be a preliminary signal of **where the G in AGI** can come from.[3]

- RESULT 6. We explain *why* mistakes occur. For instance, the model makes systematic errors that can be explained by probing its internal states. Sometimes, these mistakes can be predicted before the model generates answers, making them independent of the random generation process. We connect this to practice, noting that GPT-4/4o also makes similar errors (though we cannot probe their internal states).

- RESULT 7+8. The depth of the language model is crucial for its reasoning ability. For example, a 16-layer, 576-dim transformer solves harder problems (in reasoning length) than a 4-layer, 1920-dim one, despite the latter being twice as large. This holds even when Chain-of-Thought (CoT) is used. We explain this necessity in depth by the complexity of the mental processes involved. We advocate for the use of controlled, synthetic data as a more principled approach to derive such claims, contrasting with predictions like "only size matters" based on training loss using internet pretrain data [14].

---

[1]There is a rich literature studying how well language models can learn arithmetic and length generalization, see [13, 23] and the references therein. Modern language models are also equipped with retrieval-augmented generation (RAG), allowing arithmetic computations to be delegated to a calculator.

[2]In our case, one can solve all the math problems without computing all-pair dependency. Our pretraining data never includes such information — all the solutions only compute necessary variables.

[3]Indeed, the skill to sort relationships among in-context objects is a general skill, which may lead to — via instruction fine-tuning — skills for solving other tasks, such as discovering causal relationships, determining the influence of parameter changes, etc.

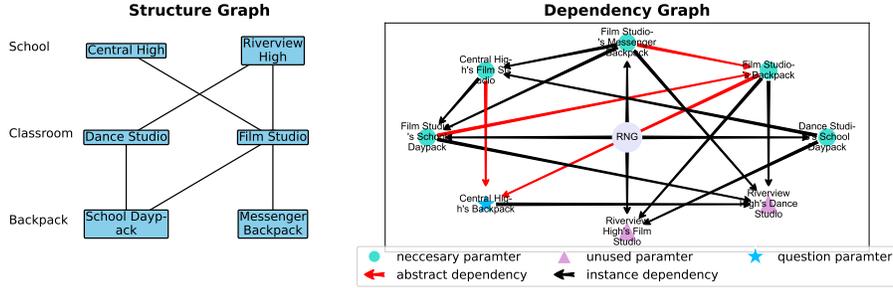

Figure 1: Structure and dependency graph corresponding to the op = 7 easy example in (2.1) and (2.2). Dependencies from abstract parameters are drawn in **red**, and from instance parameters are in **black**.

While we refrain from overstating that our findings directly apply to foundation models like GPT-4 or more challenging mathematical reasoning tasks, we believe our work significantly advances the understanding of how language models develop their mathematical reasoning skills, and this **has to be done in a way different from pushing benchmarks**.

## 2   Result 1: Data Generation

**Motivation.**   Recall a standard grade-school math problem in the GSM8K dataset [9] looks like:

> Betty is saving money for a new wallet which costs 100. Betty has only half of the money she needs. Her parents decided to give her 15 for that purpose, and her grandparents twice as much as her parents. How much more money does Betty need to buy the wallet?

This problem involves multiple parameters whose values are connected through various equalities, such as "Betty's current money = $0.5 \times$ cost of the wallet" and "money given by grandparents = $2 \times$ money given by parents." Motivated by this, we build a GSM8K-like math dataset through a synthetic generation pipeline that captures the dependencies of parameters. We wish to capture at least the following three types of dependencies.

1. Direct dependency ($\heartsuit$): such as $A = 5 \times (X + Y)$, so $A$ can be computed after $X$ and $Y$.

2. Instance dependency ($\spadesuit$): such as "every classroom has X chairs, and there are Y classrooms." Here, the model must infer the total number of chairs by multiplying X by Y.

3. Implicit dependency ($\clubsuit$): such as "Bob has 3 times more fruits than Alice. Alice has 3 apples, 4 eggs and 2 bananas." Here, the model must learn that apples and bananas are fruits and egg is not, and "Alice's fruits" is an abstract parameter derived from the problem statement.

### 2.1   Step 1: Graph Construction and Problem Generation

**Hierarchical categorization.**   We use a layered structure of <u>*categories*</u>, each contains possible <u>*items*</u>. For instance, categories = (School, Classroom, Backpack) has three layers; category School = {Central High, Riverview High, . . . }; category Classroom = {Dance Studio, Film Studio, . . . }; category Backpack = {School Daypack, Messenger Backpack, . . . }. We prepare 4 predefined hierarchical categorizations, each with 4 layers and 100 items in each layer; this represents the world knowledge.

**Structure graph.** In each math problem, only specific items exist, leading to a *structure graph* that outlines what sub-items can appear under what item, see Figure 1 (left). For instance,

- Connecting Dance Studio and School Daypack with an edge signifies an *instance parameter*, "the number of school daypacks in each dance studio," which is a quantifiable variable that can be assigned.[4] This captures the instance dependency (♠) as mentioned above.

- *Abstract parameters*, like "the total number of classrooms in Central High," cannot be assigned and are excluded from the structure graph. They reflect implicity dependency (♣).

*Remark* 2.1. Rather than using simple objects like *Alice's apple* or fake items like *Items A/B/C/D*, this structure allows us to describe abstract parameters and adds 2 levels of complexity to the data:

- The model must implicitly learn English concepts, such as a classroom category includes 100 different classroom types. These concepts cannot be derived from individual math problems, as only a limited selection of classrooms will be mentioned in each problem.

- The model is required to hierarchically access multiple items to calculate abstract parameters, as opposed to a straightforward retrieval of "Alice's apple" in the context.[5]

**Dependency graph.** The *dependency graph* is a directed acyclic graph that outlines the dependency among parameters. For each *instance parameter*, we choose a random set of (up to 4) parameters it can depend on — including possibly a special vertex RNG representing a random number generator. For instance, if "[param A] is $X$ more than the difference of [param B] and [param C]" for $X$ being randomly generated, then we draw edges from B, C and RNG to parameter A. The dependency of abstract parameters is implied by the dependency of instance parameters. This captures direct dependency (♡) as mentioned above. We give an examples on the right side of Figure 1, and details for how we randomly generate such dependency graph are in Appendix D.2.

**Problem generation.** The *problem* is articulated by describing the dependency graphs in English, one sentence for each instance parameter.[6] (Abstract parameters are not described because they are inherited by the structure graph.) We **randomly permute** the sentence ordering to further increase difficulty. A parameter is selected and asked with a question in the end (or at the beginning). Below is an easy example corresponding to Figure 1; a harder example is in Figure 11.

> **(Problem - Easy)** The number of each Riverview High's Film Studio equals 5 times as much as the sum of each Film Studio's Backpack and each Dance Studio's School Daypack. The number of each Film Studio's School Daypack equals 12 more than the sum of each Film Studio's Messenger Backpack and each Central High's Film Studio. The number of each Central High's Film Studio equals the sum of each Dance Studio's School Daypack and each Film Studio's Messenger Backpack. The number of each Riverview High's Dance Studio equals the sum of each Film Studio's Backpack, each Film Studio's Messenger Backpack, each Film Studio's School Daypack and each Central High's Backpack. The number of each Dance Studio's School Daypack equals 17. The number of each Film Studio's Messenger Backpack equals 13. *How many Backpack does Central High have?*
>
> (2.1)

---

[4]Even though Central High and Rivierside High can both have (possibly multiple) Dance Studios, for simplicity, we assume that each Dance Studio has the same number of School Daypacks.

[5]For example, the total number of backpacks in Riverview High in Figure 1 is calculated as $ip_1 \times ap_1 + ip_2 \times ap_2$ where $ip_1$ = "Riverview High's number of Dance Studios", $ip_2$ = "Riverview High's number of Film Studios", $ap_1$ = "each Dance Studio's number of Backpacks", and $ap_2$ = "each Film Studio's number of Backpacks", with $ip_1, ip_2$ being instance parameters and $ap_1, ap_2$ abstract parameters. Here, the model must not only retrieve $ip_1, ip_2$ but also compute $ap_1, ap_2$ hierarchically.

[6]We use simple English sentence templates to describe the problem, and did not worry about grammar mistakes such as singular vs plural forms. There are other randomness besides the dependency graph, such as when parameter A depends on B, C it could be $A + B$ or $A - B$.

## 2.2 Step 2: Solution Construction (CoT)

Let *solution* be a sequence of sentences describing the *necessary* steps towards solving the given problem, where the sentences follow any topological order — also known as Chain-of-Thought, CoT. For each parameter *necessary* towards answering the final question, we assign to it a random letter among the 52 choices (a..z or A..Z), and use a sentence to describe its computation:[7]

> Define [param] as X; [intermediate steps]; so X = ...

Throughout this paper, we consider **arithmetics mod 23** to avoid errors from computation involving large numbers. It is perhaps the easiest to directly see a solution example (corresponding to (2.1)), and a more involved example is in Figure 11:

> **(Solution - Easy)** Define Dance Studio's School Daypack as p; so p = 17. Define Film Studio's Messenger Backpack as W; so W = 13. Define Central High's Film Studio as B; so B = p + W = 17 + 13 = 7. Define Film Studio's School Daypack as g; R = W + B = 13 + 7 = 20; so g = 12 + R = 12 + 20 = 9. Define Film Studio's Backpack as w; so w = g + W = 9 + 13 = 22. Define Central High's Backpack as c; so c = B * w = 7 * 22 = 16. *Answer: 16.*

(2.2)

We emphasize that:
- The solution only contain parameters *necessary* towards calculating the final query parameter.
- The solution follows the correct logical order: i.e. all the parameters used in the calculation must have appeared and been computed beforehand.
- We break computations to binary ops: $g = 12+13+7$ is broken into $g = 12+R$ and $R = 13+7$ in the above solution. The number of semicolons ";" equals the number of *operations*. This reduces the arithmetic complexity of the solution, which is not the focus of this paper.[8]

## 2.3 Difficulty Control

Although deferring all the pseudocode to Appendix D, we summarize below the main randomness used in the data generation process. This includes the random choice of a hierarchical categorization (i.e., the English part); a structure graph (i.e., the instance parameters); a dependency graph; arithmetic computations on the dependency graph; integer numbers (i.e., the RNG); problem sentence permutation; and the query parameter.

We use two parameters to control data's difficulty: ip is the number of instance parameters, and op is the number of solution operations; the data's difficulty is an increasing function over them. We call our dataset iGSM, to reflect the nature that such synthetic dataset can be of *infinite size*. We use $\mathsf{iGSM}^{\mathsf{op} \leq op, \mathsf{ip} \leq ip}$ to denote the data generated with constraint $\mathsf{op} \leq op$ and $\mathsf{ip} \leq ip$, and use $\mathsf{iGSM}^{\mathsf{op} = op, \mathsf{ip} \leq ip}$ to denote those restricting to $\mathsf{op} = op$. [9]

## 2.4 Train and Test Datasets

We consider two families of datasets.
- In the iGSM-med data family we use $\mathsf{ip} \leq 20$.

  The training data is $\mathsf{iGSM\text{-}med}^{\mathsf{op} \leq 15} \stackrel{\text{def}}{=} \mathsf{iGSM}^{\mathsf{op} \leq 15, \mathsf{ip} \leq 20}$. We evaluate the pretrained model both in-distribution, on $\mathsf{iGSM\text{-}med}^{\mathsf{op} \leq 15}$ and $\mathsf{iGSM\text{-}med}^{\mathsf{op} = 15}$, and out-of-distribution (OOD), on iGSM-med$^{\text{op}=op}$ for $op \in \{20, 21, 22, 23\}$ and iGSM-med$^{\text{op}=op,\text{reask}}$. Here, reask denotes first generating a problem from iGSM-med$^{\text{op}=op}$ and then resampling a query parameter.[10]

---

[7]There are different ways to format the CoT solution. We noted that starting with "Define [param] as X" instead of [intermediate steps] improves the model's accuracy, so we have adhered to this CoT format.

[8]Even GPT-4 can make mistakes on calculating "3 * (4+10) + 12 * (5+6)" without using external calculator.

[9]We choose op non-uniformly; for instance, we let $\mathsf{op} = \min\{t_0, t_1\}$ for two random draws $t_0, t_1 \in [op]$. This ensures that the dataset has more easy data — which makes training faster. (See also similar behavior for arithmetics [13].)

| | op=2 | op=3 | op=4 | op=5 | op=6 | op=7 | op=8 | op=9 | op=10 | op=11 | op=12 | op=13 | op=14 | op=15 | op=16 | op=17 | op=18 | op=19 | op=20 |
|---|---|---|---|---|---|---|---|---|---|---|---|---|---|---|---|---|---|---|---|
| 5-shot on GPT-4-turbo | 96.7% =29/30 | 90.0% =27/30 | 80.0% =24/30 | 93.3% =28/30 | 56.7% =17/30 | 56.7% =17/30 | 50.0% =15/30 | 50.0% =15/30 | 43.3% =13/30 | 33.3% =10/30 | 40.0% =12/30 | 33.3% =10/30 | 43.3% =13/30 | 30.0% =9/30 | 36.7% =11/30 | 26.7% =8/30 | 46.7% =14/30 | 43.3% =13/30 | 40.0% =12/30 |
| 5-shot on GPT-4o | 96.7% =29/30 | 100% =30/30 | 90.0% =27/30 | 93.3% =28/30 | 86.7% =26/30 | 86.7% =26/30 | 73.3% =22/30 | 70.0% =21/30 | 70.0% =21/30 | 50.0% =15/30 | 43.3% =13/30 | 56.7% =17/30 | 46.7% =14/30 | 36.7% =11/30 | 43.3% =13/30 | 46.7% =14/30 | 36.7% =11/30 | 46.7% =14/30 | 46.7% =14/30 |
| guess ans=0 | | 23.5% | 28.8% | 29.7% | 30.7% | 30.6% | 31.4% | 30.3% | 31.9% | 30.9% | 30.1% | 31.8% | 31.2% | 33.0% | 30.4% | 32.3% | 32.6% | 32.8% | 33.8% | 34.5% |

Figure 2: GPT-4 [17] few-shot accuracies on iGSM-med$_{pq}$ (with mod5 arithmetics). For each op we tested 30 problems; and guessing $ans = 0 \in \{0, 1, 2, 3, 4\}$ gives a baseline accuracy around 32%. Details are in Appendix G, where we also give showcase how GPT-4/4o make mistakes.

- In the iGSM-hard data family we use ip $\leq 28$.
  The training data is iGSM-hard$^{\text{op}\leq 21} \stackrel{\text{def}}{=}$ iGSM$^{\text{op}\leq 21,\text{ip}\leq 28}$. We evaluate the pretrained model both in-distribution, on iGSM-hard$^{\text{op}\leq 21}$ and iGSM-hard$^{\text{op}=21}$, and OOD on iGSM-hard$^{\text{op}=op}$ for $op \in \{28, 29, 30, 31, 32\}$ and iGSM-hard$^{\text{op}=op,\text{reask}}$.

Additionally, we use iGSM-med$_{pq}$ to indicate placing the question *after* the problem and iGSM-med$_{qp}$ the other way (similarly for iGSM-hard). The difficulty of iGSM-med is already quite non-trivial to humans (at least not solvable with few-shot learning using GPT-4/4o, see Figure 2).

**Proposition 2.2.** *Ignoring unused parameters, numerics, sentence orderings, English words, a-z and A-Z letter choices, iGSM-med$^{\text{op}=15}$ still has at least* 7 *billion* <u>solution templates</u>, *and iGSM-hard$^{\text{op}=21}$ has at least* 90 *trillion* <u>solution templates</u>.[11]

**No data contamination.** A goal in synthetic math data generation is to prevent data contamination in internet-based math datasets, as noted in [22]. While it *may be impossible to certify that models trained on internet data are free from contamination*, in our setting, **we can certify this**:

1. We perform OOD evaluation such as on op $\geq 28$ while providing only op $\leq 21$ training samples.
2. We train with data whose hash value of *solution template* (see Footnote 11) is $< 17 \pmod{23}$, and test with those $\geq 17$. This ensures *no template-level overlap between training and testing*.

## 3 Result 2-3: Summarize Model's Behavior Process

We use the GPT2 architecture [18] but replacing its absolute positional embedding with rotary embedding [7, 20], yet still referring to it as GPT2 for short.[12] We mostly stick to the 12-layer, 12-head, 768-dim GPT2 (a.k.a. GPT2-small) for experiments, but we explore larger models in Section 6. We use a context length of 768 / 1024 for pretraining on iGSM-med/iGSM-hard and 2048 for evaluation. More details are in Appendix F.

---

[10]Due to the topological nature of our data/solution generation process, reask greatly changes the data distribution and the number of operations needed. It provides an excellent OOD sample for evaluation. Details are in Appendix D.

[11]A solution template is created by replacing all numbers with '0', substituting variables (a-z or A-Z) with letters in their appearance order, and changing parameters to their types (instance or abstract). For instance, "Define Owl Forest's Elephant as y; so y = 11. Define Parrot Paradise's Raccoon as t; so t = y = 11." becomes "Define Inst as a; so a = 0. Define Inst as b; so b = a = 0." We use birthday paradox to estimate the number of solution templates. If $M$ randomly generated problems yield distinct templates, it suggests with good probability that the total number of templates exceeds $\Omega(M^2)$.

[12]We also tested with Llama architecture (esp. with gated MLP layers) and did not see any benefit of using it. GPT2-rotary performs no worse than Llama/Mistral for knowledge tasks [4]. We are currently bounded by resources to repeat all experiments in this paper with other architectures that have minor differences from GPT2-rotary.

|  | iGSM-med_pq | | iGSM-med_qp | | iGSM-hard_pq | | iGSM-hard_qp | |
|---|---|---|---|---|---|---|---|---|
|  | in-dist | out-of-dist (OOD) | in-dist | out-of-dist (OOD) | in-dist | out-of-dist (OOD) | in-dist | out-of-dist (OOD) |
| beam1 - nosample | 99.9 99.1 | 91.8 87.9 84.0 76.8 91.6 | 100 99.3 | 92.4 89.9 84.8 78.2 91.4 | 100 99.4 | 94.4 92.0 90.6 86.8 82.8 91.3 | 100 99.2 | 94.5 93.2 91.0 88.2 85.3 89.4 |
| beam4 - dosample | 99.9 99.1 | 92.0 88.4 84.5 77.7 91.6 | 100 99.1 | 92.4 89.6 84.7 78.3 91.4 | 100 99.3 | 94.2 92.2 90.3 86.5 82.5 91.3 | 99.9 99.2 | 94.4 93.3 90.8 87.7 85.3 89.1 |

op≤15  op=15  op=20  op=21  op=22  op=23  op=20 (reask)   op≤15  op=15  op=20  op=21  op=22  op=23  op=20 (reask)   op≤21  op=21  op=28  op=29  op=30  op=31  op=32  op=28 (reask)   op≤21  op=21  op=28  op=29  op=30  op=31  op=32  op=28 (reask)

Figure 3: Test accuracies on the model (pre-)trained from the iGSM-med$_{pq/qp}$ and iGSM-hard$_{pq/qp}$ datasets.

|  | iGSM-med_pq | iGSM-med_qp | iGSM-hard_pq | iGSM-hard_qp |
|---|---|---|---|---|
| avg unnecessary operation | 0.00 0.00 0.00 0.00 0.00 0.00 | 0.00 0.00 0.00 0.00 0.00 0.00 | 0.01 0.00 0.00 0.00 0.00 0.00 0.00 | 0.00 0.00 0.00 0.00 0.00 0.00 0.00 |
| avg unnecessary operation (reask) | 0.02 0.11 0.15 0.17 0.19 0.17 | 0.03 0.11 0.17 0.21 0.19 0.20 | 0.07 0.46 0.52 0.54 0.57 0.66 0.66 | 0.09 0.40 0.45 0.45 0.58 0.53 0.59 |
| avg unnecessary parameter | 0.00 0.00 0.00 0.00 0.00 0.00 | 0.00 0.00 0.00 0.00 0.00 0.00 | 0.01 0.00 0.00 0.00 0.00 0.00 0.00 | 0.00 0.00 0.00 0.00 0.00 0.00 0.00 |
| avg unnecessary parameter (reask) | 0.01 0.09 0.12 0.12 0.14 0.12 | 0.03 0.10 0.15 0.16 0.15 0.16 | 0.06 0.34 0.36 0.36 0.38 0.43 0.44 | 0.07 0.30 0.32 0.31 0.40 0.36 0.42 |

op≤15  op=15  op=20  op=21  op=22  op=23    op≤15  op=15  op=20  op=21  op=22  op=23    op≤21  op=21  op=28  op=29  op=30  op=31  op=32    op≤21  op=21  op=28  op=29  op=30  op=31  op=32

Figure 4: Number of unnecessary params / operations used per generated correct solution. Details in Appendix F.

**Result 2: accuracy.** After sufficient pre-training, we give the model a problem from the test set (without solution) and let it continue to generate (allegedly a solution followed by an answer). Because we have restricted ourselves to a fixed solution format, language models can learn the format easily, allowing us to write a *solution parser* to check if the solution is fully correct.[13]

> **Result 2.** *Figure 3 shows that GPT2 performs well when pretrained using iGSM-med or iGSM-hard data, even when evaluated out-of-distribution on harder (i.e., larger op) math problems. Thus, the model can truly learn some reasoning skill instead of memorizing solution templates.*[14]

This could be reminiscent of language models' length generalization capability on arithmetic computations [13, 23]; however, in our case, op captures the "reasoning length" in grade-school math, and our model **has never seen *any*** training example of the same length as in test time.[15]

Such accuracies also indicate that our iGSM data families are indeed good for pretraining purpose, allowing us to investigate how LLMs can solve grade-school math problems.

**Result 3: solution redundancy.** We examine whether GPT2 achieves high accuracy by

- brute-forcedly computing all the parameters during generation (a "level-0" reasoning skill), or
- computing only necessary parameters to give shortest solutions (a "level-1" reasoning skill).

Recall our iGSM (pretrain) data only contains necessary solution steps (i.e., CoT) to simulate what we see in textbook solutions for math problems. For instance, if a problem describes X =3+2, E =3+X, Y =X+2 and asks for the value of Y, then a shortest solution would be "X =3+2=5 and Y =X+2 =7" without ever computing E.

> **Result 3.** *Figure 4 shows that GPT2 predominantly solves the iGSM problems with a "level-1" reasoning skill, avoiding unnecessary computations, even when evaluated out-of-distribution.*

This finding is significant as it suggests that, unlike humans who usually rely on "backward reasoning" and a scratch pad to write down necessary parameters by backtracking the dependencies from the question [19], the language model can directly generate shortest solutions without using a scratch pad. But, how does it achieve so? We shall investigate in the next section.

---

[13]We check not only the correctness of the final answer 0..22 but also the calculations and parameter dependencies. Language models can learn very complex syntactics, see [1] and the references therein.

[14]Llama (of the same model size) gives similar performance, but we refrain from repeating all the experiments with another model. We are not interested in small model differences in this theoretical study; instead, we care more about the general behavior of (autoregressive) language models.

[15]Some others such as Anil et al. [6] start with a transformer pre-trained on internet data; while the transformer may not have seen the same task during training, it's possible that the model has seen other tasks with the same (or even longer) length and learned to transfer from there.

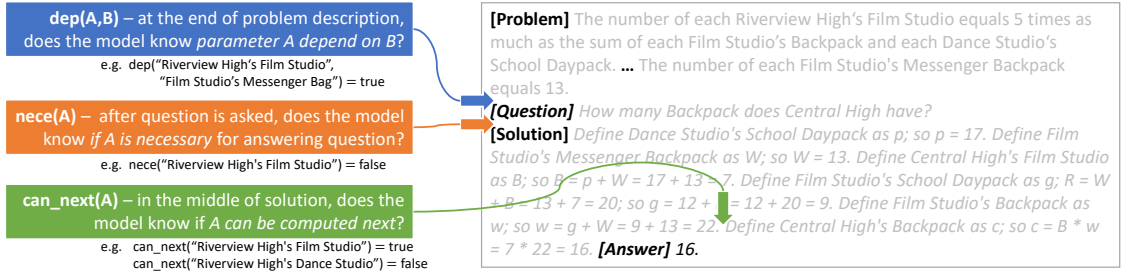

Figure 5: To discover model's mental (reasoning) process.

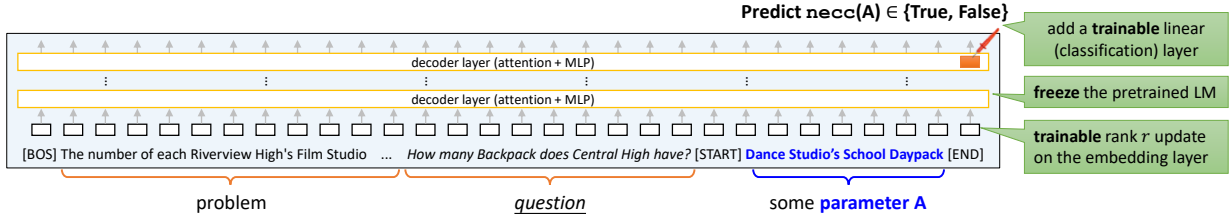

Figure 6: Illustrations of V-probing on the `nece`($A$) task. For other tasks, see Figure 13.

## 4 Result 4-5: Discover Model's Mental Process

To understand how the model learns to solve math problems, we propose studying the following probing tasks, which align closely with human problem-solving strategies:
- `nece`($A$): if parameter $A$ is necessary for computing the answer.
- `dep`($A, B$): if parameter $A$ (recursively) depends on parameter $B$ given the problem statement.
- `known`($A$): if parameter $A$ has already been computed.
- `value`($A$): the value of parameter $A$ (a number between 0-22, or 23 if `known`($A$) = false).
- `can_next`($A$): if $A$ can be computed in the next solution sentence (namely, its predecessors have all been calculated). Note that $A$ might not be necessary to answer the question.
- `nece_next`($A$): if parameter $A$ satisfies both `can_next`($A$) and `nece`($A$).

For a model to generate the shortest solutions, it must identify `nece`($A$) for all $A$'s in its mental process. This is because whether `nece`($A$) is true directly corresponds to whether there is a solution sentence to compute $A$. However, *how early* does the model recognize this, and how is it stored? Similarly, does it recognize dependencies between parameters (`dep`)? If so, how early is this mental process completed? Moreover, in the *middle of solution generation*, does the model keep track of each parameter $A$'s value at all times (`value`, `known`)? Does the model mentally know all possible parameters $A$ that are ready to compute in the next sentence (`can_next`)? Or does it only focus on $A$ that is both ready and necessary (`nece_next`)?

This section proposes probing technique to answer all of these questions.

Figure 7: V-probing accuracies (for beam=1; results for beam=4 are almost identical). Details are in Appendix F.2.

(a) Probing accuracies on the six tasks: can_next(A), dep(A, B), known(A), nece(A), nece_next(A), value(A).

(b) Probing accuracies of can_next(A), dep(A, B) *restricted* to positives/negatives labels in which A is unnecessary

## 4.1 V-Probing: A Nearly-Linear Probing Method

As illustrated in Figure 5, we conduct probing at the end of the problem description for the dep task, and end of the question description nece task.[16] For other tasks, we probe them at the end of *every* solution sentence (including the start of the first solution sentence).

Recall that standard *linear probing* involves freezing a pretrained language model and checking if a property is linearly encoded at a hidden layer (usually the last layer) for a given token position. This is done by introducing a trainable linear classifier on the hidden states and performing a lightweight finetuning task for this property (see [12] and references therein).

Our setting is more complex because the properties have one or two conditional variables, $A$ and $B$, described in plain English. To handle this, we truncate the math problems to the probing position and append tokens [START] and [END] around the descriptions of $A$ (or $A, B$). We then probe from the token position of [END] to see if the property is linearly encoded at the last layer.

Unlike standard linear probing, to account for the input change, we introduce a small trainable rank-8 (linear) update on the input embedding layer. We freeze the pretrained language model and finetune both the linear classifier and the rank-8 update for the desired property. We refer to this as V(ariable)-probing and provide details in Appendix B. An illustration of the nece(A) probing task is shown in Figure 6.

We compute the V-probing accuracies on a language model pretrained from iGSM and compare them with the V-probing accuracies on a randomly-initialized transformer model. If the former accuracies are significantly higher, we conclude that the probing signals must have (or be very close to having) come from the pretrained weights, rather than the (lightweight) finetuning stage.

## 4.2 Probing Results and Findings

We present our probing results in Figure 7. The probing accuracies are high for all the tasks, compared to majority guess and random-model probing — except for the very hard OOD cases

---

[16]If the problem format is qp (question asked before the problem) then we probe nece and dep both after the problem description.

(i.e., for large op where the model's generation accuracies fall down to 80% anyways in Figure 3),

**Result 4: model solves math problems like humans.** We make the following observations:

- When generating solutions, the model not only remembers which parameters have been computed and which have not (`value`, `known`) but also knows which parameters can be computed next (`can_next`, `nece_next`). These abilities ensure that the model can solve the given math problem step by step, similar to human problem-solving skills.

- By the end of the problem description, the model already knows the full list of necessary parameters (`nece`). This indicates that the model has learned to *plan ahead*, identifying necessary parameters before starting to generate the solution. This aligns with human behavior, except that the model plans mentally while humans typically write this down. This further confirms that the model reaches the "level-1" reasoning skill discussed in Section 3.

*Remark* 4.1. The mental process described can be compared to (out-of-context) *knowledge manipulation* [2], which involves retrieving factual knowledge and performing single-step computations (e.g., retrieving two people's birth dates to determine who was born earlier). Allen-Zhu and Li [2] found that even single-step computations *cannot* be performed mentally without a substantial number of pretrain samples. In contrast, this paper studies *in-context* reasoning and demonstrates that the model can execute very complex mental calculations.

**Result 5: model learns beyond human reasoning skills.** Remarkably, the model learns $\text{dep}(A, B)$ and $\text{can\_next}(A)$, even for parameters $A$ not necessary for answering the question, as shown in Figure 7(b). This differs from human problem-solving, where we typically use backward reasoning from the question to identify necessary parameters, often overlooking unnecessary ones [19]. In contrast, language models can pre-compute the all-pair dependency graph $\text{dep}(A, B)$ mentally even before a question is asked. We consider this a "level-2" reasoning skill that is very different from human behavior or mental processes.

Thus, although this skill is not needed for solving the math problems and although no pretrain data teaches the model to compute "all-pair dependency" — fitting the data only requires computing necessary parameters — the model still discovers it after training. This enables the model to sort relationships among the things it hears, a skill that can be useful for future tasks (via instruction fine-tuning). To our knowledge, this may be the first evidence of a language model acquiring skills *beyond* those needed for learning its pretrain data; and this may be a preliminary signal of where the G in AGI can come from (generalizing to skills not taught in the pretrain data).

**Corollary: the backward thinking process.** A key question for AGI success is whether the "backward thinking process" (e.g., "because I want to compute X, but X depends on Y and Y depends on Z, so let me compute Z first") needs to be explicitly included in the training data. This differs from CoT, where CoT breaks down complex computations into simpler steps, but planning is still required to decide which step to compute first.

Our findings suggest that, at least for grade-school math problems, with abundant data, this backward thinking process can be autonomously learned through language modeling, without needing to be directly included in the training data.

## 5 Result 6: Explain Model's Mistakes

We further examine the relationship between our probing results and the model's generated solutions, focusing on two questions: (1) When does the model answer correctly but include unnecessary

|  | nece(A) | | | | | | | | | |
|---|---|---|---|---|---|---|---|---|---|---|
|  | iGSM-med | | | | | iGSM-hard | | | | |
| on all parameters \| pq | 99.8% | 98.7% | 97.9% | 96.9% | 94.7% | 99.6% | 99.1% | 98.6% | 97.9% | 97.1% | 95.5% |
| on all parameters \| pq (reask) | 93.4% | 94.9% | 95.7% | 95.6% | 95.9% | 89.6% | 89.9% | 90.5% | 91.6% | 92.0% | 92.0% |
| on all parameters \| qp | 99.9% | 99.5% | 99.4% | 99.3% | 99.2% | 99.8% | 99.7% | 99.7% | 99.6% | 99.4% | 99.3% |
| on all parameters \| qp (reask) | 98.5% | 98.1% | 98.3% | 98.5% | 98.2% | 96.6% | 96.1% | 96.6% | 97.1% | 97.2% | 97.2% |
| on unnecessary parameter in model's output \| pq (reask) \| beam1 | 31.6% =105/332 | 38.6% =168/435 | 49.2% =224/455 | 40.5% =206/509 | 35.9% =161/449 | 20.5% =262/1280 | 17.1% =225/1314 | 18.6% =245/1320 | 21.6% =298/1381 | 23.2% =364/1570 | 23.6% =370/1566 |
| on unnecessary parameter in model's output \| pq (reask) \| beam4 | 30.3% =92/304 | 38.1% =167/438 | 45.8% =204/445 | 40.5% =217/536 | 34.7% =169/487 | 19.7% =252/1280 | 16.5% =219/1326 | 18.4% =242/1316 | 20.1% =272/1353 | 23.3% =355/1523 | 23.1% =346/1496 |
| on unnecessary parameter in model's output \| qp (reask) \| beam1 | 21.7% =81/373 | 22.5% =124/551 | 24.0% =145/605 | 21.6% =118/546 | 21.7% =125/575 | 17.0% =193/1133 | 22.3% =260/1165 | 21.1% =234/1108 | 21.0% =303/1443 | 18.9% =243/1285 | 24.7% =367/1484 |
| on unnecessary parameter in model's output \| qp (reask) \| beam4 | 21.8% =80/367 | 22.5% =126/560 | 22.6% =133/588 | 21.1% =120/569 | 20.6% =123/597 | 17.1% =194/1136 | 22.6% =254/1125 | 23.1% =251/1085 | 21.6% =304/1405 | 19.0% =252/1325 | 25.0% =375/1498 |
|  | op=15 | op=20 | op=21 | op=22 | op=23 | op=21 | op=28 | op=29 | op=30 | op=31 | op=32 |

(a) nece($A$) probing accuracies correlate with model's outputted unnecessary parameters

|  | can_next(A) | | | | | | | | nece_next(A) | | | | | | | |
|---|---|---|---|---|---|---|---|---|---|---|---|---|---|---|---|---|
|  | iGSM-med | | | | iGSM-hard | | | | | iGSM-med | | | | iGSM-hard | | | |
| on all parameters \| pq | 99.3% | 99.2% | 99.1% | 99.0% | 99.1% | 99.0% | 99.0% | 98.9% | 98.9% | 99.2% | 99.1% | 99.0% | 98.7% | 98.7% | 98.6% | 98.4% | 98.3% | 98.3% |
| on all parameters \| qp | 98.8% | 98.6% | 98.7% | 98.5% | 99.1% | 99.0% | 99.0% | 99.0% | 99.0% | 98.7% | 98.3% | 98.3% | 97.7% | 99.5% | 99.4% | 99.3% | 99.2% | 99.2% |
| on first wrong param \| pq \| beam1 | 75.8% =172/227 | 75.6% =270/357 | 69.6% =330/474 | 70.9% =473/667 | 57.1% =128/224 | 58.2% =185/318 | 60.7% =229/377 | 63.7% =332/521 | 62.5% =419/670 | 52.2% =119/228 | 47.6% =171/359 | 41.6% =197/474 | 42.5% =284/669 | 42.9% =96/224 | 40.9% =130/318 | 45.4% =171/377 | 46.4% =242/522 | 47.2% =317/671 |
| on first wrong param \| pq \| beam4 | 76.6% =187/244 | 76.3% =280/367 | 70.4% =350/497 | 69.8% =481/689 | 61.3% =141/230 | 59.9% =187/312 | 62.9% =246/391 | 66.8% =356/533 | 63.3% =439/694 | 51.6% =126/244 | 47.2% =174/369 | 44.9% =223/497 | 43.3% =300/693 | 39.6% =91/230 | 40.3% =126/313 | 45.5% =178/391 | 46.4% =248/534 | 48.8% =339/695 |
| on first wrong param \| qp \| beam1 | 68.1% =190/279 | 65.2% =234/359 | 66.5% =354/532 | 67.8% =503/742 | 59.2% =119/201 | 59.4% =149/251 | 62.4% =204/327 | 60.5% =259/428 | 61.9% =313/506 | 51.3% =143/279 | 56.1% =202/360 | 56.5% =301/533 | 59.4% =443/746 | 50.5% =102/202 | 50.6% =127/251 | 52.9% =173/327 | 52.8% =226/428 | 57.3% =290/506 |
| on first wrong param \| qp \| beam4 | 67.9% =190/280 | 67.7% =249/368 | 65.7% =352/536 | 69.7% =524/752 | 63.0% =131/208 | 58.5% =148/253 | 64.1% =214/334 | 57.8% =256/443 | 60.9% =310/509 | 54.6% =153/280 | 60.4% =223/369 | 56.3% =302/536 | 60.6% =458/756 | 54.5% =114/209 | 51.4% =130/253 | 56.3% =188/334 | 54.9% =243/443 | 56.0% =285/509 |
|  | op=20 | op=21 | op=22 | op=23 | op=28 | op=29 | op=30 | op=31 | op=32 | op=20 | op=21 | op=22 | op=23 | op=28 | op=29 | op=30 | op=31 | op=32 |

(b) can_next($A$) and nece_next($A$) probing accuracies correlate with model's outputted wrong solutions

Figure 8: Probing results correlate with model's output solutions. We tested 4096 math problems and presented the probing accuracies restricted to (1) unnecessary parameters in the model's correct output solution (**top**), and (2) the first wrong parameter in model's wrong output solution (**bottom**). Details are in Appendix F.2.

parameters? (2) What causes incorrect answers? We aim to determine if such erroneous behavior of the model aligns with errors in the model's mental process (via probing).

For the first question, given the model rarely produces solutions longer than necessary (see Figure 4), we turned to out-of-distribution reask data for evaluation.[17] On this data, pretrained models produce an average of $\sim 0.5$ unnecessary parameters per solution even for op = 32 (see Figure 4). We examined if these unnecessary parameters $A$ were **incorrectly predicted as** nece($A$) = true in the probing task. Figure 8(a) reveals that this is often indeed the case, thus language models produce solutions with unnecessary steps due to errors in their *mental planning phase*.

For the second question, we focused on the model's *wrong* solutions and their *first wrong parameters*. (Using synthetic data, we can easily identify such parameters.) Our findings in Figure 8(b) show that the model's errors mainly stem from **incorrectly predicting nece_next($A$) or can_next($A$) as true** in its internal states when such $A$'s are not ready for computation.[18]

> **Result 6** (Figure 8). *Combining these, we conclude:*
> - *Many reasoning mistakes made by the language model are systematic, stemming from* errors in its mental process, *not merely random from the generation process.*
> - *Some of the model's mistakes can be discovered by probing its inner states* even before the model opens its mouth *(i.e., before it says the first solution step).*

We also observe that GPT-4/4o makes similar mistakes by outputting unnecessary parameters

---

[17] Recall this re-samples a query after generating the problem, leading to a different set of necessary parameters.

[18] In Figure 8(b), we focus on these "first wrong parameters" with correct label being can_next($A$) = false or nece_next($A$) = false and present the probability that their probing also correctly predicts false. Low accuracy indicates that the model "thought" these parameters were ready for computation, but they were not.

|                          | iGSM-med_pq |                       | iGSM-med_qp |                       | iGSM-hard_pq |                       | iGSM-hard_qp |                       | avg  |
|--------------------------|:-----------:|:---------------------:|:-----------:|:---------------------:|:------------:|:---------------------:|:------------:|:---------------------:|:----:|
|                          | in-dist     | out-of-dist (OOD)     | in-dist     | out-of-dist (OOD)     | in-dist      | out-of-dist (OOD)     | in-dist      | out-of-dist (OOD)     |      |
| dep4 - size1 - head21    | 99.5 92.7   | 74.7 68.0 62.4 54.5   | 99.4 93.3   | 73.3 66.8 61.1 54.6   | 98.9 90.8    | 72.4 67.7 62.1 57.1 50.6 | 99.1 89.8 | 69.4 62.2 57.8 52.3 45.7 | 72.2 |
| dep4 - size2 - head30    | 99.6 94.7   | 74.2 67.9 61.6 53.1   | 99.4 94.5   | 78.1 71.9 65.7 58.8   | 97.0 71.7    | 46.3 40.6 37.0 32.3 27.3 | 99.4 92.1 | 74.5 69.5 64.7 59.1 53.2 | 68.6 |
| dep8 - size1 - head15    | 100 98.8    | 89.7 86.5 82.8 76.8   | 100 99.2    | 92.4 88.5 84.2 78.7   | 100 99.1     | 94.6 92.0 89.7 86.4 82.2 | 100 99.0  | 92.2 89.6 86.2 82.4 77.3 | 90.3 |
| dep8 - size2 - head21    | 100 99.3    | 93.7 91.6 88.3 83.6   | 99.9 99.0   | 90.2 87.1 83.3 76.3   | 100 99.2     | 93.6 91.3 88.6 85.6 82.6 | 100 99.1  | 93.5 91.3 89.1 85.7 81.2 | 91.3 |
| dep12 - size1 - head12   | 100 99.3    | 92.0 88.9 84.2 77.9   | 100 99.4    | 92.2 89.2 83.9 77.9   | 100 99.5     | 96.0 94.1 91.0 88.5 84.3 | 100 99.3  | 95.3 93.0 91.9 88.0 84.5 | 91.9 |
| dep12 - size2 - head17   | 100 99.5    | 94.0 91.9 89.0 82.7   | 100 99.0    | 90.8 85.4 80.2 73.2   | 100 99.8     | 97.1 95.5 93.5 91.8 88.0 | 100 99.5  | 94.5 91.9 88.9 86.8 81.3 | 92.1 |
| dep16 - size1 - head10   | 100 99.6    | 94.6 91.9 87.9 82.7   | 100 99.5    | 89.9 85.0 79.1 71.1   | 100 99.6     | 97.0 95.2 94.2 92.2 88.5 | 100 99.4  | 95.8 93.8 92.4 88.9 85.8 | 92.5 |
| dep16 - size2 - head15   | 100 99.8    | 95.9 93.7 90.4 86.5   | 100 99.8    | 95.6 93.5 90.3 84.3   | 100 99.7     | 97.5 96.3 95.1 92.9 89.5 | 100 99.8  | 97.3 96.0 94.2 91.9 88.9 | 95.0 |
| dep20 - size1 - head9    | 100 99.8    | 95.5 93.6 90.0 86.3   | 100 99.6    | 94.8 91.4 87.4 80.4   | 100 99.8     | 97.0 95.1 94.0 91.0 87.4 | 100 99.6  | 96.6 94.5 92.8 90.1 86.5 | 94.0 |
| dep20 - size2 - head13   | 100 99.8    | 95.8 93.3 89.2 84.4   | 100 99.6    | 93.7 91.8 87.4 81.3   | 100 99.8     | 98.0 96.7 95.9 93.9 90.9 | 100 99.9  | 97.5 96.0 95.2 92.4 89.7 | 94.7 |

Column groups: op≤15, op=15, op=20, op=21, op=22, op=23 | op≤15, op=15, op=20, op=21, op=22, op=23 | op≤21, op=21, op=28, op=29, op=30, op=31, op=32 | op≤21, op=21, op=28, op=29, op=30, op=31, op=32

Figure 9: Accuracies for GPT2 models of different depth/widths pretrained on iGSM datasets. Details in Appendix F.

or insisting on computing parameters $A$ with can_next$(A)$ = false (see Appendix G). This further hints that our findings may be applicable more broadly.

## 6  Result 7-8: Depth vs. Reasoning Length

Our controlled dataset enables a systematic exploration of the relationship between a language model's depth and its reasoning length.

Recent studies have demonstrated that for knowledge storage and extraction, only model size matters (even for 2-layer transformers) [4]. Furthermore, both the seminal scaling-law paper by OpenAI [14] and theoretical studies in deep learning [5] suggest that model depth/width might have a minimal impact universally. Contrary to these findings, we present evidence that[19]

> **Result 7** (Figure 9). *Language model depth is crucial for mathematical reasoning.*

Specifically, we experimented with models of depths 4/8/12/16/20 and two sizes (a smaller size 1 and a larger size 2).[20] From Figure 9, we observe that a 4-layer transformer, even with 1920 hidden dimensions, underperforms on our math datasets. Conversely, deeper but smaller models, such as a 20-layer 576-dim, perform very well. Comparing accuracies vertically reveals a clear correlation between model depth and performance. Thus, we infer that depth is likely essential for reasoning tasks, such as solving grade-school math problems.

Next, we try to reveal "why" this happens. We delved into how depth influences math problem-solving skills through the nece probing task, focusing on necessary parameters at distance $t$ from the query parameter, for $t \in \{1, 2, \ldots, 8\}$. These parameters all have nece$(A)$ = true, but we can probe the model to see how correct they are at predicting nece$(A)$ at different hidden layers.

Figure 10 shows our result. It reveals a correlation between the model's layer hierarchy, reasoning accuracy, and *mental reasoning depth*. Shallower layers excel at predicting nece$(A)$ for parameters $A$ closer to the query, whereas deeper layers are more accurate and can predict nece$(A)$ for parameters further from the query. This suggests that the model employs layer-by-layer reasoning during the planning phase to recursively identify all parameters the query depends on, and:

> **Result 8** (Figure 10+14). *The depth of a language model is crucial, likely due to the complexity of its hidden (mental) reasoning processes. A $t$-step mental reasoning, such as mentally computing nece$(A)$ for parameters $A$ that are a distance $t$ from the query, may require deeper models for larger $t$, assuming all other hyperparameters remain constant.*

---

[19]Math reasoning data only occupies a tiny fraction of pretraining data for language models, thus one might not observe a difference if we only look at the perplexity as in the original scaling law paper [14].

[20]GPT2-$\ell$-$h$ represents an $\ell$-layer, $h$-head, $64h$-dimensional GPT2 model. Size-1 models are GPT2-4-21, GPT2-8-15, GPT2-12-12, GPT2-16-10, GPT2-20-9, with similar parameter counts; size-2 models are GPT2-4-30, GPT2-8-21, GPT2-12-17, GPT2-16-15, GPT2-20-13, approximately twice the size of size-1 models.

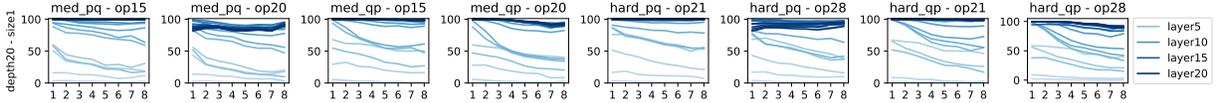

Figure 10: Increasing probing accuracies of `nece`($A$) with increasing layer depth. The x-axis denotes the distance of parameter $A$ to the query parameter, with **colors from light to dark to represent layers 1 to 20**. This figure is for a 20-layer GPT2 model; for other model depths/sizes, see Figure 14.

We make two disclaimers here. First, if the "backward thinking process" is added as CoT to the data (see the end of Section 4.2), then deep mental thinking is no longer required, reducing the language model's depth requirement. However, in practice, many such "thinking processes" may not be included in standard math solutions or languages in general.

Second, the above claim does not imply that "a $t$-step mental thinking requires a depth-$t$ transformer". It is plausible for a single transformer layer (containing many sub-layers) to implement $t > 1$ mental thinking steps, though possibly with reduced accuracy as $t$ increases. We refrain from providing an exact correlation in this paper, as it heavily depends on the data distribution.

# 7 Conclusion

We use a synthetic setting to demonstrate that language models can learn to solve grade-school math problems through true generalization, rather than relying on data contamination or template memorization. We develop probing techniques to examine the models' hidden reasoning processes. Our findings reveal that these models can learn math skills aligned with human cognitive processes, as well as "new thinking processes" not present in the training data. Additionally, we propose a method to predict a model's errors before it begins to solve a problem and to explain why models make mistakes when they occur. Based on this discovery, we write a separate paper to improve language models' math reasoning accuracy [21]. We also provide a principled approach to connect the model's depth to its capable reasoning length. We believe this research opens doors to study the mathematical reasoning skills of language models from a different angle compared to pushing math benchmarks.

One may argue that iGSM may be very different from the pretrain data that modern LLMs use. While this may be true, we are looking into the future. Recall, even GPT-4/4o of today cannot few-shot learn to solve iGSM-med$^{\text{op}=11}$ (see Figure 2). From this perspective, it is reasonable to believe that future versions of LLMs will rely on synthetic math pretrain data to improve their reasoning skills. While one may not directly use iGSM, it is tempting to use existing LLMs (such as Llama-3) to turn iGSM into more natural formats while keeping the logical chains. On the other hand, we have discovered that models trained purely on the iGSM data make similar mistakes compared to GPT-4/4o (see Section 5 and Appendix G). This further confirms that our findings do connect to practice, regarding the model's hidden reasoning process.

Finally, Part 2 of this work series focuses on how language models solve grade-school math problems (including Part 2.2 [21]). We also cover how language models learn language structures in Part 1 [1] (in particular, how they mentally perform dynamical programming), and learn world knowledge in Part 3 [2–4].

# Appendix

## A  Result 1 — An Example in iGSM-hard with op = 21

**(Problem- A Hard Example)** The number of each Jungle Jim's International Market's Cheese equals the sum of each Parmesan Cheese's Pear and each The Fresh Market's Ice Cream. The number of each Ice Cream's Pineapple equals 2 more than each Goat Cheese's Grape. The number of each New Seasons Market's Goat Cheese equals the sum of each Residential College District's Jungle Jim's International Market, each Jungle Jim's International Market's Parmesan Cheese and each Residential College District's Supermarket. The number of each Arts Campus's New Seasons Market equals each Cheese's Pineapple. The number of each Goat Cheese's Banana equals each Vocational School District's Product. The number of each Residential College District's Jungle Jim's International Market equals 5 more than each Ice Cream's Grape. The number of each Parmesan Cheese's Pineapple equals each Parmesan Cheese's Pear. The number of each Residential College District's The Fresh Market equals each Arts Campus's Trader Joe's. The number of each Arts Campus's Trader Joe's equals each Parmesan Cheese's Ingredient. The number of each Goat Cheese's Grape equals 0. The number of each The Fresh Market's Ice Cream equals 13 more than the difference of each Residential College District's The Fresh Market and each Parmesan Cheese's Grape. The number of each Goat Cheese's Pineapple equals each New Seasons Market's Product. The number of each Vocational School District's The Fresh Market equals the sum of each Trader Joe's's Cheese and each The Fresh Market's Cheese. The number of each Trader Joe's's Cheese equals 6. The number of each The Fresh Market's Cheese equals 3. The number of each Jungle Jim's International Market's Ice Cream equals the difference of each Ice Cream's Banana and each Goat Cheese's Grape. The number of each Jungle Jim's International Market's Parmesan Cheese equals each Ice Cream's Pineapple. The number of each Parmesan Cheese's Pear equals the difference of each Goat Cheese's Grape and each Ice Cream's Grape. The number of each Parmesan Cheese's Grape equals 12 times as much as each Residential College District's Jungle Jim's International Market. The number of each The Fresh Market's Parmesan Cheese equals each The Fresh Market's Cheese. The number of each Ice Cream's Banana equals the sum of each Parmesan Cheese's Pineapple and each Ice Cream's Pineapple. The number of each School District's Jungle Jim's International Market equals each The Fresh Market's Ice Cream. The number of each Cheese's Pineapple equals 20 more than the sum of each Trader Joe's's Cheese and each The Fresh Market's Cheese. The number of each Trader Joe's's Parmesan Cheese equals 16. The number of each Ice Cream's Pear equals 8. The number of each Ice Cream's Grape equals each Goat Cheese's Grape. *How many Product does School District have?*

**(Solution- A Hard Example)** Define Goat Cheese's Grape as u; so u = 0. Define Ice Cream's Grape as x; so x = u = 0. Define Residential College District's Jungle Jim's International Market as N; so N = 5 + x = 5 + 0 = 5. Define Parmesan Cheese's Pear as G; so G = u- x = 0 - 0 = 0. Define Parmesan Cheese's Grape as f; so f = 12 * N = 12 * 5 = 14. Define Parmesan Cheese's Pineapple as C; so C = G = 0. Define Parmesan Cheese's Ingredient as Z; e = f + C = 14 + 0 = 14; so Z = e + G = 14 + 0 = 14. Define Arts Campus's Trader Joe's as q; so q = Z = 14. Define Residential College District's The Fresh Market as j; so j = q = 14. Define Ice Cream's Pineapple as X; so X = 2 + u = 2 + 0 = 2. Define Ice Cream's Banana as K; so K = C + X = 0 + 2 = 2. Define The Fresh Market's Ice Cream as P; i = j- f = 14 - 14 = 0; so P = 13 + i = 13 + 0 = 13. Define Jungle Jim's International Market's Ice Cream as R; so R = K- u = 2 - 0 = 2. Define School District's Jungle Jim's International Market as V; so V = P = 13. Define Jungle Jim's International Market's Cheese as v; so v = G + P = 0 + 13 = 13. Define Jungle Jim's International Market's Parmesan Cheese as S; so S = X = 2. Define Jungle Jim's International Market's Product as y; U = S + R = 2 + 2 = 4; so y = U + v = 4 + 13 = 17. Define School District's Product as J; so J = V * y = 13 * 17 = 14. *Answer: 14.*

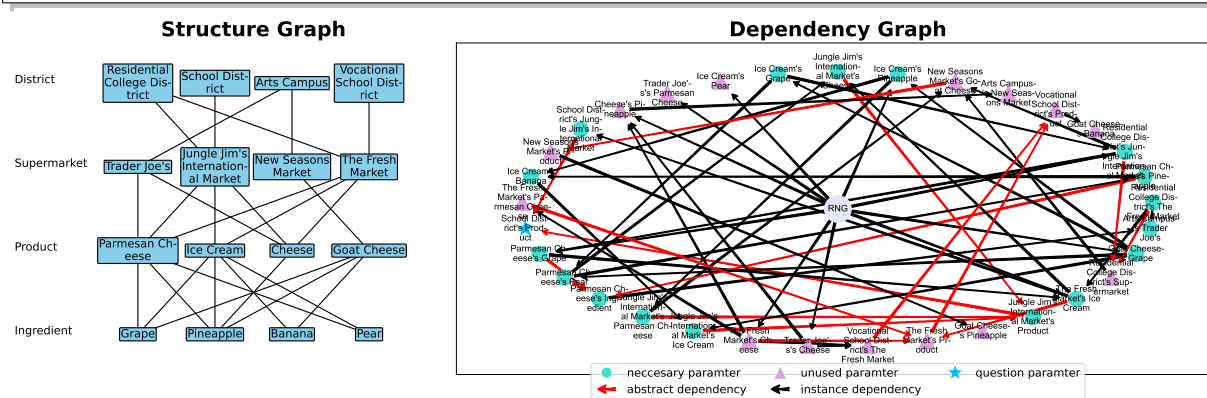

Figure 11: An example with op = 21 in iGSM-hard$_{pq}$ used for training. Don't forget during testing we evaluate models on op = 28 which is even harder.

Figure 12: Probing accuracies *restricted* to positives/negatives labels (complement to Figure 7 which is on all labels.)

## B  Results 4-5 — Details on V-probing

Recall that we wish to conduct probing at the end of the problem description for the `nece` and `dep` tasks (before the solution for `nece`; before the solution or even the question for `dep`). For other tasks, we probe at the end of *every* solution sentence (including the start of the first solution sentence). The goal is to *freeze* a pretrained language model, then introduce a *very small* number of additional trainable parameters on top of it, and finetune them for each probing task.

Specifically, we take a pretrained language model, e.g., pretrained from the iGSM-hard training data. We *freeze* its parameters completely except for adding a trainable rank-$r$ update on the embedding layer to account for the task change (from next-token prediction to probing). Throughout this paper we use a small value $r = 8$. We feed this network with training data that are the same as iGSM-hard, but truncated at exactly the position we wish to probe. Importantly, we append such inputs with a special starting token [START] along with a parameter name (or two names, if it is the $\text{dep}(A, B)$ task). We then extract the hidden states of the last token position at the last transformer layer, and add a trainable linear layer (a.k.a. linear head) to perform classification for one of the six probing tasks.

This probing method is illustrated in Figure 13. We call it V(ariable)-Probing, because it can take an arbitrary number of variables (i.e., parameters in this paper) to allow us to perform functional probing inside the transformer.

Note, if it were only a trainable linear head such probing would be called linear probing [12]. Unlike traditional linear probing, we are adding a small low-rank update on the model's embedding layer. This is arguably the minimum change needed (to account for the task change, for special tokens like [START] [MID] [END], etc.) in order to perform any non-trivial probing. This is related but different from the nearly-linear probing methods introduced in Allen-Zhu and Li [1, 3], because they do not support taking variables as probing inputs.[21]

**Unbalanced probing tasks.** Our probing accuracies for the six tasks were presented in Figure 7. However, we notice that the `dep` and `nece_next` tasks have unbalanced labels — even guessing "all false" would give 83% accuracy for $\text{dep}(A, B)$ and 92% for $\text{nece\_next}(A)$. For such reason, we also present their probing accuracies restricted to positives/negatives labels separately in Figure 12.

---

[21]In Allen-Zhu and Li [1, 3], the authors are interested in probing the model's behavior via fixed classification tasks (such as a 100-class classification task) given data that are identical or nearly-identical to the pretrain data. In this paper, we are interested in the model's behavior with respect to given variables (such as parameter names, which can have $\sim 100k$ possibilities); and we append such variable names to the input to make the training inputs appear very different from the original pretrain data.

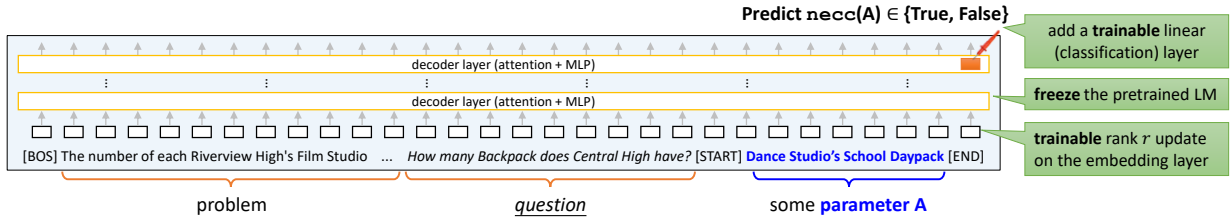
(a) V-probing for the $\texttt{nece}(A)$ task

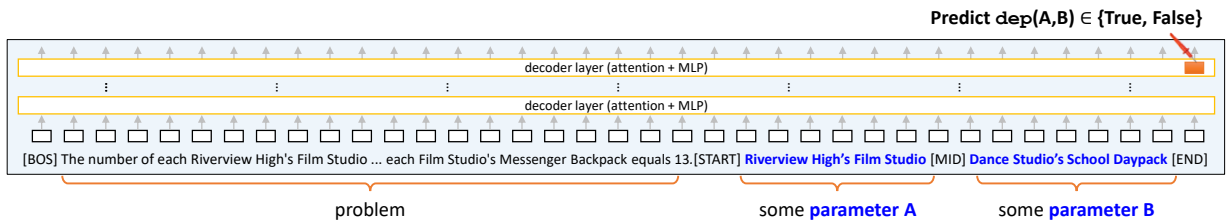
(b) V-probing for the $\texttt{dep}(A, B)$ task

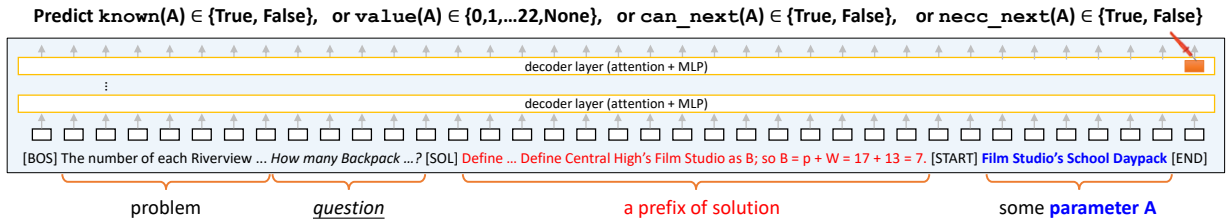
(c) V-probing for the $\texttt{value}(A), \texttt{can\_next}(A), \texttt{nece\_next}(A)$ tasks

Figure 13: Illustrations of V-probing, our nearly-linear probing methods to investigate whether a pretrained model, at a specific input position, *knows* an arbitrary $\texttt{func}(A)$ for a parameter $A$ described in text.

In all cases, we freeze the entire pretrained language model, except for a low-rank $r = 8$ update on the input embedding layer to accommodate the task change.

The illustration is for pq data (problem precedes question); for qp data, we simply reverse the order, except for $\texttt{dep}(A, B)$ where the question is added before the problem.

# C  Result 8 — Additional Figure

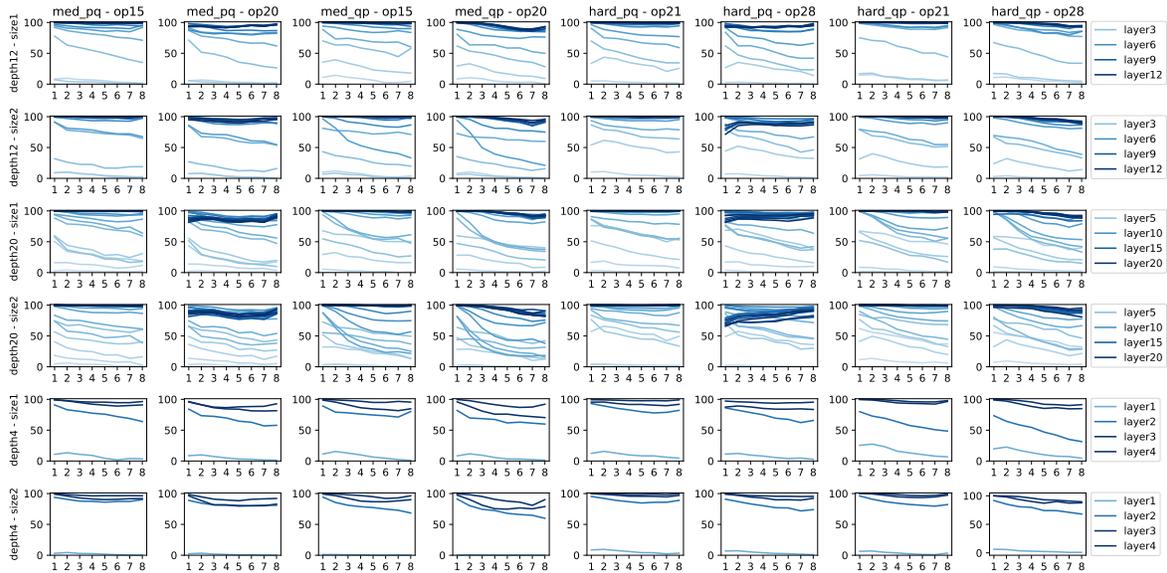

Figure 14: Increasing probing accuracies of nece($A$) with increasing layer depth. This is an extension of Figure 10 but including more model depths/sizes.
The x-axis denotes the distance of parameter $A$ from the query parameter, with colors transitioning from light to dark to represent layers 1 to max. (Model architecture details are in Footnote 20 and Appendix F.)

# D Result 1 Details — Math Data Generation

Our math data generation process consists of first generating the structure graph (see Figure 1 and 11 left), which defines the set of parameters we shall use; then generating the dependency graph (see Figure 1 and 11 right), which defines the arithmetic relationship between the parameters; and finally generating the English problem and solution descriptions.

**Notations.** In this section, to make the description concise, when we say "randomly sampling" in the pseudocode, we mean uniform random unless otherwise noted. Whenever we consider a (directed) graph $G$, slightly abusing notation, we write $a \in G$ to indicate that $a$ is a vertex in $G$ and $(a \to b) \in \mathsf{G}$ to indicate that there is an edge from $a$ to $b$ in $G$.

## D.1 Generate Structure Graph

Recall the structure graph (see Figure 1 and 11 left) describes the set of possible items (nodes) and instance parameter (edges) that we shall rely on to construct our math problem.

We use $G_\mathsf{s}$ to denote such structure graph, and it is generated $G_\mathsf{s} = \texttt{DrawStructure}(e, d, w_0, w_1)$ from a random distribution defined with *hyperparameters* $e, d, w_0, w_1 \in \mathbb{N}$. At a high level, we construct $G_\mathsf{s}$ so that it has $d$ layers, $e$ edges, and each layer has between $w_0$ and $w_1$ items.

Specifically, suppose $l_i \in \{w_0, w_0+1, \ldots, w_1\}$ represents the number of items for each layer $i$. In this configuration, one must have at least $e^- = l_2 + \cdots + l_d$ edges to ensure the graph is "connected", and at most $e^+ = l_1 l_2 + \cdots + l_{d-1} l_d$ edges. Using this formula, we first randomly choose a configuration $(l_1, \ldots, l_d)$ so that $e^- \leq e \leq e^+$ for the given parameter $e$. Then, after the configuration is chosen, we randomly generate edges accordingly. Details are given in Algorithm 1.

---

**Algorithm 1** $G_\mathsf{s} = \texttt{DrawStructure}(e, d, w_0, w_1)$
---
**Input:** $e, d, w_0, w_1 \in \mathbb{N}$ ⋄ *satisfying* $2 \leq d \leq 4$; $2 \leq w_0 \leq w_1 \leq 4$; $(d-1)w_0 \leq e \leq (d-1)w_1^2$
1: $l \leftarrow (w_0, w_0, \ldots, w_0) \in \mathbb{Z}^d$ ⋄ $l_i$ *represents the number of items (nodes) for layer $i$*
2: $p \leftarrow$ uniform random from $(0, 1)$
3: **while** $l \neq (w_1, w_1, \ldots, w_1)$ **do**
4:   $e^-, e^+ \leftarrow$ minimum and maximum number of edges that $l$ can give
5:   **if** $e^+ < e$ **then**
6:     randomly select $i \in [d]$ such that $l_i < w_1$, and increase it $l_i \leftarrow l_i + 1$.
7:   **else if** $e^- = e$ **then**
8:     **break**
9:   **else if** randomly choose a number in $(0, 1)$ and it is less than $p$ **then**
10:     randomly select $i \in [d]$ such that $l_i < w_1$, and increase it $l_i \leftarrow l_i + 1$.
11:   **else**
12:     **break**
13: **end** ⋄ *after while loop, we must have $e^- \leq e \leq e^+$ and $\forall i \in [d]$: $w_0 \leq l_i \leq w_1$*
14: Construct $G_\mathsf{s}$ with exactly $l_i$ items on layer $i \in [d]$.
15: **for** each item $a$ in each layer $i \geq 2$ **do**
16:   randomly select an item $b$ in layer $i - 1$ and connect $(a, b)$ in $G_\mathsf{s}$. ⋄ *this creates $e^-$ edges*
17: **while** number of edges $< e$ **do**
18:   randomly select two items $a, b$ from adjacent layers to create an edge in $G_\mathsf{s}$.
19: **return** $G_\mathsf{s}$ and attach English to it.

### D.1.1 Attach English

As described in Section 2.1, we have prepared 4 predefined hierarchical categorizations, each of them with 4 total layers of categories:

```
[
    ["District", "Supermarket", "Product", "Ingredient"],
    ["Zoo", "Enclosure", "Animal", "Bone"],
    ["School", "Classroom", "Backpack", "Stationery"],
    ["Ecosystems", "Creatures", "Organs", "Cells"]
]
```

In each of the above 16 categories, we have prepared around 100 items (further decomposed into 5 sub-categories). Below is a showcase of them:

```
{
    "District": {
        "Residential Districts": [...],
        "Commercial Districts": [
            "Shopping District", "Business District", "Financial District", "
                                                Industrial District",
            "Warehouse District", "Market District", "Restaurant District", "
                                                Entertainment District",
            "Arts District", "Fashion District", "Silicon Valley", "Wall Street",
            "Tech Park", "Automotive District", "Jewelry District", "Medical
                                                District",
            "Legal District", "Media District", "Research Park", "Manufacturing
                                                District"
        ],
        "Historical Districts": [...],
        "Educational Districts": [...],
        "Government Districts": [...]
    },
    "Supermarket": {...},
    "Product": {
        "Canned Foods": [...],
        "Snack Foods": [
            "Potato Chips", "Pretzels", "Popcorn", "Candy Bars",
            "Gummy Candy", "Cookies", "Crackers", "Granola Bars",
            "Fruit Snacks", "Cheese Puffs", "Nuts", "Trail Mix",
            "Beef Jerky", "Rice Cakes", "Yogurt Covered Raisins", "Chocolate
                                                Covered Pretzels",
            "Tortilla Chips", "Salsa", "Hummus", "Dried Fruit"
        ],
        "Beverages": [...],
        "Baked Goods": [...],
        "Dairy Products": [...]
    },
    "Ingredient": {...},
    "Zoo": {...},
    "Enclosure": {...},
    "Animal": {...},
    "Bone": {...},
    "School": {...},
    "Classroom": {...},
    "Backpack": {...},
    "Stationery": {...},
    "Ecosystems": {...},
    "Creatures": {...},
```

```
      "Organs": {...},
      "Cells": {...}
}
```

Now, given a constructed structure graph $G_s$, we first randomly pick one of the four categorizations, then randomly pick $d \in \{2, 3, 4\}$ consecutive layers of categories, next randomly pick one of the five subcategories, and finally pick $l_i$ random item names in this subcategory for each layer $i$.

At this point, we have constructed $G_s$ as well as added English names to each of its node, just like Figure 1 and 11 (left).

## D.2 Generate Dependency Graph

A structure graph $G_s$ defines the set of possible parameters we consider, while a *dependency graph* defines how these parameters depend on each other. We use an edge $a \to b$ to indicate that parameter $b$ depends on $a$; there is a special vertex RNG and it can happen that $\mathsf{RNG} \to b$. What an abstract parameter depends on is inherited from the structure graph $G_s$. For each instance parameter, we shall randomly add edges to indicate what parameters it depends on.

**High-level plan.** We shall use $G_d$ to denote the dependency graph, we start from an empty graph and then add vertices/edges incrementally and randomly. Our process is as follows:

- Generate a *necessary dependency graph* $G_d^{\mathsf{nece}}$ which covers all the vertices and nodes that are necessary for the computation of the query parameter.

  - Generate necessary abstract parameters (and add parameters they depend on); call this graph $G_d^{\mathsf{nece1}}$.
  - Generate necessary instance parameters and add them to $G_d^{\mathsf{nece1}}$; call this graph $G_d^{\mathsf{nece2}}$.
  - Generate a topological order for parameters $G_d^{\mathsf{nece2}}$ and ensure all of them are necessary towards computing the query parameter (which is the last one in this tropologic order). During this process, we shall add additional edges from $G_d^{\mathsf{nece2}}$ to create $G_d^{\mathsf{nece3}}$.
  - Generate additional necessary edges and add them to $G_d^{\mathsf{nece3}}$; call this graph $G_d^{\mathsf{nece}}$.

- Add to $G_d^{\mathsf{nece}}$ all the remaining (unnecessary) parameters and edges to form $G_d$.

At a high level, our problem description shall solely depend on $G_d$— by describing each instance parameter in it using a sentence, and our solution description shall solely depend on $G_d^{\mathsf{nece}}$— by describing the computation of each parameter in it using a sentence.

Before we proceed with the construction let us formally introduce:

**Definition D.1** (operation). *Given any dependency graph $G_d$,*

- *For an (abstract or instance) parameter $a \in G_d$ that has in-degree $t \geq 0$, we define $\mathsf{op}_{G_d}(a) \stackrel{\text{def}}{=} \max\{1, t-1\}$ which is the number of* operations *needed to compute $a$.*[22]

- *We use $\mathsf{op}(G_d) \stackrel{\text{def}}{=} \sum_{a \in G_d \setminus \{\mathsf{RNG}\}} \mathsf{op}_{G_d}(a)$ to denote the total number of (arithmetic) operations needed to compute all the parameters in $G_d$.*

*Remark* D.2. In our final design of $G_d$, we shall ensure that each parameter (except the special vertex RNG) has in-degree at least 1; however, during the construction process since we add edges incrementally, some (instance) parameter may temporarily have in-degree 0. For notation simplicity, we still say $\mathsf{op}_{G_d}(a) = \max\{1, -1\} = 1$ in such a case.

---

[22]For instance, in Figure 1, $a =$ "Riverview High's total number of Backpacks" is equal to $ip_1 \times ap_1 + ip_2 \times ap_2$ for $ip_1 =$ "Riverview High's number of Dance Studios", $ip_2 =$ "Riverview High's number of Film Studios", $ap_1 =$ "each Dance Studio's number of Backpacks", $ap_2 =$ "each Film Studio' number of Backpacks", where $ip_1, ip_2$ are instance parameters and $ap_1, ap_2$ are abstract parameters. In this case, this abstract parameter depends on 4 other parameters, and requires 3 arithmetic operations.

**Hyperparameters.** We use hyperparameters $1 \leq n \leq m \leq s$ to control the difficulty of $G_d$.

- we shall ensure $\mathsf{op}(G_d^{\mathsf{nece1}}) \leq n$ and is as close as possible to $n$;
- we shall ensure $\mathsf{op}(G_d^{\mathsf{nece3}}) = \mathsf{op}(G_d^{\mathsf{nece2}}) \leq m$ and is as close as possible to $m$;
- we shall ensure $\mathsf{op}(G_d^{\mathsf{nece}}) = s$ is exact.

In other words, hyperparameter $s$ controls exactly how many operations are needed to compute the query parameter, which is the primary factor controlling the problem's difficulty.

### D.2.1 Construction of $G_d^{\mathsf{nece1}}, G_d^{\mathsf{nece2}}$

Given a structure graph $G_s$, recall its edges represent all the instance parameters we shall use. Its abstract parameters are those ones that describe quantities across 1 or multiple layers: for instance in Figure 1, Central High's number of Classrooms is across 1 layer, and Central High's number of Backpacks is across 2 layers. We define this number as the *difficulty level* of abstract parameters.

With this notion, our construction of $G_d^{\mathsf{nece1}}$ and $G_d^{\mathsf{nece2}}$ are described together in Algorithm 2.

At a high level, we try to incrementally and randomly add abstract parameters to $G_d^{\mathsf{nece1}}$ while maintaining $\mathsf{op}(G_d^{\mathsf{nece1}}) \leq n$. We cannot make this exact equality because when adding a single abstract parameter requires also (recursively) adding all the other parameters it may depend on. We tried to prioritize adding abstract parameters with higher difficulty levels. Once we finish constructing $G_d^{\mathsf{nece1}}$, we randomly add additional instance parameters from $G_s$ to make it $G_d^{\mathsf{nece2}}$.

---

**Algorithm 2** $G_d^{\mathsf{nece2}} = \texttt{DrawNecessary1}(G_s, n, m)$

**Input:** structure graph $G_s$ of depth $d$, $n, m \in \mathbb{N}$ with $1 \leq n \leq m$
1: $G_d^{\mathsf{nece1}} \leftarrow$ empty graph
2: **repeat**
3:     updated $\leftarrow$ false
4:     **for** $i \leftarrow d-1, \ldots, 1$ **do**
5:         **if** $\exists$ abstract parameter of difficulty level $i$ in $G_s$ that is not yet in $G_d^{\mathsf{nece1}}$ **then**
6:             randomly pick one such abstract parameter $a$ of difficulty level $i$
7:             $G' \leftarrow G_d^{\mathsf{nece1}} + a$ and all instance/abstract parameters $a$ may (recursively) depend on
                                                                                                  $\diamond$ *also add their dependency edges*
8:             **if** $\mathsf{op}(G') \leq n$ **then**
9:                 $G_d^{\mathsf{nece1}} \leftarrow G'$; updated $\leftarrow$ true; **break**
10: **until** updated $=$ false
11: $G_d^{\mathsf{nece2}} \leftarrow G_d^{\mathsf{nece1}}$     $\diamond$ $\mathsf{op}(G_d^{\mathsf{nece1}}) \leq n$ *and all instance parameters in* $G_d^{\mathsf{nece1}}$ *have in-degree 0*
12: **for** $i \leftarrow 1, 2, \ldots, m - \mathsf{op}(G_d^{\mathsf{nece1}})$ **do**
13:     if there's leftover instance parameter in $G_s$ not yet in $G_d^{\mathsf{nece2}}$, add a random one to $G_d^{\mathsf{nece2}}$
14: **return** $G_d^{\mathsf{nece2}}$     $\diamond$ $\mathsf{op}(G_d^{\mathsf{nece2}}) \leq m$ *and all instance parameters in* $G_d^{\mathsf{nece2}}$ *have in-degree 0*

---

### D.2.2 Construction of $G_d^{\mathsf{nece3}}$

Our goal next is to select a random query parameter in $G_d^{\mathsf{nece2}}$ and construct a random topological ordering Topo for all the parameters in $G_d^{\mathsf{nece2}}$, so as to ensure that all the parameters are necessary towards the computation of query.

We start with `Topo = [query]` and append parameters to its left one by one. During this process, we may also introduce new edges randomly; we start with $G_{\mathsf{d}}^{\mathsf{nece3}} = G_{\mathsf{d}}^{\mathsf{nece2}}$ and add edges incrementally. This process may not always succeed — sometimes the created topological ordering cannot make all the parameters necessary towards the computation of the `query`. If this happens we declare a failure.[23]

We introduce two notions (we use $G_{\mathsf{d}}^{\mathsf{nece3}} \setminus \texttt{Topo}$ to denote the set of vertices in $G_{\mathsf{d}}^{\mathsf{nece3}}$ that are not in `Topo`):

- $\mathsf{Next1}_{G_{\mathsf{d}}^{\mathsf{nece3}}}(\texttt{Topo}) \stackrel{\text{def}}{=} \{a \in G_{\mathsf{d}}^{\mathsf{nece3}} \setminus \texttt{Topo} \mid \exists (a \to b) \in G_{\mathsf{d}}^{\mathsf{nece3}} \text{ for some } b \in \texttt{Topo}\}$

  Intuitively, if $a \notin \mathsf{Next1}(\texttt{Topo})$ then we cannot immediately append $a$ to the front of `Topo`, because it is not yet necessary towards the computation of `query`.

- $\mathsf{Next2}_{G_{\mathsf{d}}^{\mathsf{nece3}}}(\texttt{Topo}) \stackrel{\text{def}}{=} \{a \in G_{\mathsf{d}}^{\mathsf{nece3}} \setminus \texttt{Topo} \mid \nexists (a \to b) \in G_{\mathsf{d}}^{\mathsf{nece3}} \text{ for any } b \in G_{\mathsf{d}}^{\mathsf{nece3}} \setminus \texttt{Topo}\}$

  Intuitively, if $a \notin \mathsf{Next2}_{G_{\mathsf{d}}^{\mathsf{nece3}}}(\texttt{Topo})$ then we cannot immediately append $a$ to the front of `Topo`, because some other parameter depends on it and is not yet added to `Topo`. (Obviously we always have $\mathsf{Next2}_{G_{\mathsf{d}}^{\mathsf{nece3}}}(\texttt{Topo}) \neq \varnothing$ unless $G_{\mathsf{d}}^{\mathsf{nece3}} \setminus \texttt{Topo} = \varnothing$ so we are done.)

Our generation algorithm is now easy to describe: we keep adding parameters that are in $\mathsf{Next1}_{G_{\mathsf{d}}^{\mathsf{nece3}}}(\texttt{Topo}) \cap \mathsf{Next2}_{G_{\mathsf{d}}^{\mathsf{nece3}}}(\texttt{Topo})$ to the front of `Topo`; and if we get stuck, we introduce new edges to $G_{\mathsf{d}}^{\mathsf{nece3}}$ (or declare failure). The pseudocode is in Algorithm 3.

---
**Algorithm 3** $(G_{\mathsf{d}}^{\mathsf{nece3}}, \texttt{Topo}) = \texttt{DrawNecessary2}(G_{\mathsf{d}}^{\mathsf{nece2}})$
---
1: $G_{\mathsf{d}}^{\mathsf{nece3}} \leftarrow G_{\mathsf{d}}^{\mathsf{nece2}}$; $\texttt{Topo} \leftarrow []$.
2: **while** true **do**
3:    **if** $\texttt{Topo} = []$ **then**
4:       $\texttt{param}_0 \leftarrow$ random parameter in $\mathsf{Next2}_{G_{\mathsf{d}}^{\mathsf{nece3}}}(\texttt{Topo})$;                 ◇ *this is* `query` *parameter*
5:    **else**
6:       $\texttt{param}_0 \leftarrow$ random parameter in $\mathsf{Next1}_{G_{\mathsf{d}}^{\mathsf{nece3}}}(\texttt{Topo}) \cap \mathsf{Next2}_{G_{\mathsf{d}}^{\mathsf{nece3}}}(\texttt{Topo})$;
7:    $\texttt{Topo} = [\texttt{param}_0] + \texttt{Topo}$                                          ◇ *append to the front*
8:    **if** $G_{\mathsf{d}}^{\mathsf{nece3}} \setminus \texttt{Topo} = \varnothing$ **then break**
9:    **if** $\mathsf{Next1}_{G_{\mathsf{d}}^{\mathsf{nece3}}}(\texttt{Topo}) \cap \mathsf{Next2}_{G_{\mathsf{d}}^{\mathsf{nece3}}}(\texttt{Topo}) = \varnothing$ **then**
10:       **If** $\texttt{param}_0$ is abstract **then return** failure
11:       $\texttt{param}_1 \leftarrow$ a "random" parameter in $\mathsf{Next2}_{G_{\mathsf{d}}^{\mathsf{nece3}}}(\texttt{Topo})$.         ◇ *see Remark D.4*
12:       add edge $\texttt{param}_1 \to \texttt{param}_0$ to $G_{\mathsf{d}}^{\mathsf{nece3}}$.           ◇ *now* $\texttt{param}_1 \in \mathsf{Next1}_{G_{\mathsf{d}}^{\mathsf{nece3}}}(\texttt{Topo})$
13:    **else if** $\texttt{param}_0$ is instance parameter **then**
14:       **if** a probability event $p_0$ occurs for $p_0$ uniform chosen in $(0, 1)$ **then**
15:          $\texttt{param}_1 \leftarrow$ a "random" parameter in $G_{\mathsf{d}}^{\mathsf{nece3}} \setminus \texttt{Topo}$.           ◇ *see Remark D.4*
16:          add edge $\texttt{param}_1 \to \texttt{param}_0$ to $G_{\mathsf{d}}^{\mathsf{nece3}}$.          ◇ *now* $\texttt{param}_1 \in \mathsf{Next1}_{G_{\mathsf{d}}^{\mathsf{nece3}}}(\texttt{Topo})$
17: **return** $(G_{\mathsf{d}}^{\mathsf{nece3}}, \texttt{Topo})$     ◇ $\mathsf{op}(G_{\mathsf{d}}^{\mathsf{nece3}}) \leq m$ *and all instance parameters in* $G_{\mathsf{d}}^{\mathsf{nece3}}$ *have in-degree* $\leq 1$
---

**Proposition D.3.** *Every instance parameter in* $G_{\mathsf{d}}^{\mathsf{nece3}}$ *has in-degree* $\leq 1$ *and thus* $\mathsf{op}(G_{\mathsf{d}}^{\mathsf{nece3}}) = \mathsf{op}(G_{\mathsf{d}}^{\mathsf{nece2}})$.

*Remark* D.4. In Line 11 and Line 15 of Algorithm 3, when randomly selecting $\texttt{param}_1$ from a set, instead of doing so uniformly at random, to improve the algorithm's success rate and the problem's difficulty level, we introduce a discursion that that biases slightly towards abstract parameters and parameters already in $\texttt{Next1}_{G_{\mathsf{d}}^{\mathsf{nece3}}}(\texttt{Topo})$.[24] Specifically, we first generate $g \sim \mathcal{N}(0, 1)$ a random Gaussian, then define $\texttt{weight}(a) = \left(\mathbb{1}_{a \text{ is abstract}} + \mathbb{1}_{a \in \texttt{Next1}_{G_{\mathsf{d}}^{\mathsf{nece3}}}(\texttt{Topo})}\right) \cdot |g|$, and then sample $a$ with a probability $\propto e^{\texttt{weight}(a)}$.

---
[23]The outside pseudocode, which comes later, shall go back to regenerate the structure graph and start again.

### D.2.3 Construction of $G_{\mathsf{d}}^{\mathsf{nece}}$

So far we have created $G_{\mathsf{d}}^{\mathsf{nece3}}$ and $\texttt{Topo}$ with the property that every instance parameter in $G_{\mathsf{d}}^{\mathsf{nece3}}$ has in-degree $\leq 1$. In the next step, we add additional dependency edges to make in-degree to be a random number between 1 and 4. We do so by introducing additional edges; and we also introduce an additional vertex RNG. This is our final *necessary* dependency graph $G_{\mathsf{d}}^{\mathsf{nece}}$.

Our pseudocode is given in Algorithm 4. In this step, we shall make sure $\mathsf{op}(G_{\mathsf{d}}^{\mathsf{nece}}) = s$ is exact (and declare failure if this is not possible). We do so to precisely control the solution's difficulty (so that when we evaluate the model, we can choose to evaluate it on problems with a fixed value of $s$).

---

**Algorithm 4** $G_{\mathsf{d}}^{\mathsf{nece}} = \texttt{DrawNecessary3}(G_{\mathsf{d}}^{\mathsf{nece3}}, \texttt{Topo}, s)$

1: $\texttt{cur\_op}(a) \leftarrow \mathsf{op}_{G_{\mathsf{d}}^{\mathsf{nece3}}}(a)$ for every parameter $a \in G_{\mathsf{d}}^{\mathsf{nece3}}$.
2: $\texttt{max\_op}_{\texttt{Topo}}(a) \stackrel{\text{def}}{=}$ the maximum number of operations an instance parameter $a$ can require.[25]
3: **while** $\sum_{a \in G_{\mathsf{d}}^{\mathsf{nece3}}} \texttt{cur\_op}(a) < s$ **do**
4: $\quad$ randomly select an instance parameter $a \in G_{\mathsf{d}}^{\mathsf{nece3}}$ with $\texttt{cur\_op}(a) < \texttt{max\_op}_{\texttt{Topo}}(a)$;
5: $\quad$ **If** $a$ is found **then** $\texttt{cur\_op}(a) \leftarrow \texttt{cur\_op}(a) + 1$ **else return** failure.
6: $G_{\mathsf{d}}^{\mathsf{nece}} \leftarrow G_{\mathsf{d}}^{\mathsf{nece3}}$ + vertex RNG.
7: **for** each instance parameter $a$ in $G_{\mathsf{d}}^{\mathsf{nece3}}$ **do**
8: $\quad$ $\texttt{pool} \leftarrow \texttt{RNG}$ + all parameters in front of $a$ in $\texttt{Topo}$.
9: $\quad$ **if** $\texttt{cur\_op}(a) = 1$ **then**
10: $\quad\quad$ $\texttt{dep\_num} \leftarrow 1$ or $2$ each w.p. $0.5$;
11: $\quad$ **else**
12: $\quad\quad$ $\texttt{dep\_num} \leftarrow \texttt{cur\_op}(a) + 1$
13: $\quad$ $\texttt{dep\_num} \leftarrow \min\{|\texttt{pool}|, \texttt{dep\_num}\}$
14: $\quad$ **if** $\exists (b \to a) \in G_{\mathsf{d}}^{\mathsf{nece3}}$ for some $b \in \texttt{pool}$ **then** $\quad\quad\quad\quad\quad\quad\quad \diamond$ *at most one such $b$*
15: $\quad\quad$ $\texttt{pool} \leftarrow \texttt{pool} \setminus \{b\}$ and $\texttt{dep\_num} \leftarrow \texttt{dep\_num} - 1$
16: $\quad$ **if** $\texttt{dep\_num} = |\texttt{pool}|$ **then**
17: $\quad\quad$ add $b \to a$ to $G_{\mathsf{d}}^{\mathsf{nece}}$ for all $b \in \texttt{pool}$;
18: $\quad$ **else**
19: $\quad\quad$ with probability $0.5$, add $\texttt{RNG} \to a$ to $G_{\mathsf{d}}^{\mathsf{nece}}$ and $\texttt{dep\_num} \leftarrow \texttt{dep\_num} - 1$
20: $\quad\quad$ $\texttt{pool} \leftarrow \texttt{pool} \setminus \{\texttt{RNG}\}$
21: $\quad\quad$ add $b \to a$ to $G_{\mathsf{d}}^{\mathsf{nece}}$ for $\texttt{dep\_num}$ randomly select elements $b$ in $\texttt{pool}$.
22: **return** $G_{\mathsf{d}}^{\mathsf{nece}}$ $\quad\quad\quad\quad\quad\quad\quad\quad\quad\quad\quad\quad\quad\quad\quad\quad\quad\quad\quad\quad\quad\quad \diamond$ $\mathsf{op}(G_{\mathsf{d}}^{\mathsf{nece}}) = s$ *is exact*

---

[24] For those who are interested, abstract parameters are the keys to cause the generation process to fail, because once they become $\texttt{param}_0$ we cannot add edges $\texttt{param}_1 \to \texttt{param}_0$; so we had better select them earlier than later (thus put them at the back of $\texttt{Topo}$). On the other hand, for $\texttt{param}_1$ that is already in $\texttt{Next1}_{G_{\mathsf{d}}^{\mathsf{nece3}}}(\texttt{Topo})$, adding this edge $\texttt{param}_1 \to \texttt{param}_0$ does not further change it; this can help us create a problem whose solution "depth" is higher.

[25] If an instance parameter $a$ is the $i$-th element in $\texttt{Topo}$, then $\texttt{max\_op}(a) = \min\{3, \max\{1, i-1\}\}$. (Recall we require each instance parameter to depend on at most 4 vertices in the dependency graph and this amounts to no more than 3 operations.)

### D.2.4 Construction of $G_{\mathsf{d}}$

Finally, once we have $G_{\mathsf{d}}^{\mathsf{nece}}$ the necessary dependency graph, we are left to add unnecessary dependency edges (and unnecessary parameters) to form the complete $G_{\mathsf{d}}$.

During this process, we shall add all the *remaining* instance parameters from $G_{\mathsf{s}}$ into $G_{\mathsf{d}}$. When adding each of them, we randomly select the parameters that it shall depend on from all the previously known parameters.[26] Note that during this process, we may also introduce new, unnecessary abstract parameters, see the full pseudocode in Algorithm 5.

*Remark* D.5. $G_{\mathsf{d}}$ consists of all the instance and query parameters in $G_{\mathsf{s}}$ and the abstract parameters they may (recursively) depend on. There may exist abstract parameters that can be described in $G_{\mathsf{s}}$ that are not present in $G_{\mathsf{d}}$; but all the instance parameters in $G_{\mathsf{s}}$ shall be present in $G_{\mathsf{d}}$.

---

**Algorithm 5** $G_{\mathsf{d}} = \mathtt{DrawUnnecessary}(G_{\mathsf{s}}, G_{\mathsf{d}}^{\mathsf{nece}})$

1: $\mathtt{IndList} \leftarrow \varnothing$;
2: **while** $\exists$ instance parameter in $G_{\mathsf{s}}$ not yet in $G_{\mathsf{d}}$ **do**
3:     $K \leftarrow$ all params in $G_{\mathsf{d}}$ + all abstract params computable using parameters in $G_{\mathsf{d}}$;
4:     randomly select an instance parameter $a$ in $G_{\mathsf{s}}$ not yet in $G_{\mathsf{d}}$; and add $a$ to $G_{\mathsf{d}}$;
5:     **if** with half probability **then**
6:        $\mathtt{pool} \leftarrow \mathtt{IndList} \cup \{\mathsf{RNG}\}$; $\mathtt{IndList} \leftarrow \mathtt{IndList} \cup \{a\}$;
7:     **else**
8:        $\mathtt{pool} \leftarrow K \cup \{\mathsf{RNG}\}$;
9:     $\mathtt{dep\_num} \leftarrow 1$
10:    **while** $\mathtt{dep\_num} < \min\{4, |\mathtt{pool}|\}$ **do**
11:       with 0.5 probability, $\mathtt{dep\_num} \leftarrow \mathtt{dep\_num} + 1$; otherwise **break**
12:    **if** $\mathtt{dep\_num} = |\mathtt{pool}|$ **then**
13:       $\mathtt{selected} \leftarrow \mathtt{pool}$
14:    **else**
15:       $\mathtt{selected} \leftarrow \{\}$
16:       with probability 0.5, add $\mathtt{selected} = \{\mathsf{RNG}\}$ and $\mathtt{dep\_num} \leftarrow \mathtt{dep\_num} - 1$
17:       $\mathtt{pool} \leftarrow \mathtt{pool} \setminus \{\mathsf{RNG}\}$
18:       $\mathtt{selected} \leftarrow \mathtt{selected} \cup \mathtt{dep\_num}$ random elements from $\mathtt{pool}$
19:    **for** each $b \in \mathtt{selected}$ **do**
20:       If $b \notin G_{\mathsf{d}}$ then recursively add $b$ and its dependencies to $G_{\mathsf{d}}$;
21:       Add $b \to a$ to $G_{\mathsf{d}}$.
22: **return** $G_{\mathsf{d}}$

---

### D.3 Generate English: Problem, Question and Solution

At this point, we have constructed a dependency graph $G_{\mathsf{s}}$ where each instance parameter $a \in G_{\mathsf{s}}$ may depend on between 1 and 4 other vertices (could be abstract, instance parameters or $\mathsf{RNG}$). We have not yet introduced how $a$ should be computed, and we do this using a random process $\mathtt{GenSentence}(G_{\mathsf{d}}, a)$ in Algorithm 6.

---

[26]In fact, we do slightly smarter than the most naive approach. If one simply lets each newly added unnecessary parameter to depend, randomly among all the parameters that have already been added to $G_{\mathsf{d}}$, then those unnecessary parameters will likely appear towards the end of the topological order. For such reason, we give it 0.5 probability to depend only on a set $\mathtt{IndList}$, which consists of newly-added, unnecessary parmaeters, that do not depend on $G_{\mathsf{d}}$. This way, the unnecessary parameters can also appear to the front of the tropologic order.

**Algorithm 6** GenSentence($G_\mathsf{d}, a$)

1: $str \leftarrow$ "The number of [name of $a$] equals"
2: $\texttt{pool} \leftarrow \{b \in G_\mathsf{d} : \exists (b \to a) \in G_\mathsf{d}\}$.
3: **if** $\mathsf{RNG} \in \texttt{pool}$ **then**
4:     $str \leftarrow str +$ " [random int between 0 and 22]"; and $\texttt{pool} \leftarrow \texttt{pool} \setminus \{\mathsf{RNG}\}$
5:     If $|\texttt{pool}| > 0$, $str \leftarrow str +$ " more than" or " times" each with probability 0.5.
6: **if** $|\texttt{pool}| = 1$ **then**
7:     $str \leftarrow str +$ " [name of $b$]" for $\texttt{pool} = \{b\}$.
8: **else if** $|\texttt{pool}| = |\{b, c\}| = 2$ **then**
9:     $str \leftarrow str +$ " the sum of [b] and [c]" or " the difference of [b] and [c]" each w.p. 0.5.
10: **else**
11:     $str \leftarrow str +$ " the sum of .., .., and .." with a random order of all elements from $\texttt{pool}$.

**Problem description.** The problem description simply consists of listing over all *instance* parameters $a \in G_\mathsf{d}$ and call GenSentence($G_\mathsf{d}, a$). We then randomly shuffle the sentences to make the problem hard. Please note the descriptions of abstract parameters are *not present* in the problem description, because they are inherited from the hierarchical categorization. This is our attempt to make our math data also capture some English meaning, that is the model also needs to learn what items are in each category, and which category is above another category, etc. This is some knowledge that cannot be learned by reading one problem — it must be learned after reading sufficiently many data.

**Question description.** Our query parameter can be either an instance or abstract parameter, and it is the last element in Topo. We use a single sentence to ask for its value "How many... does... have?" and we put this question either at the front or at the end of the problem description (depending on the data type).

**Solution description.** We generate the solution text, by going over all the (instance or abstract) parameters in Topo *in its correct order*, and generate a single sentence to compute each parameter. This process is straightforward but notationally heavy, we describe it below by examples.

- Given any instance parameter $a \in$ Topo, suppose for instance $a$ is 7 times the sum of parameters $b, c, d$. Because of the topological order, the parameters $b, c, d$ must have already defined with variable names, denoted as $var_b, var_c, var_d$. Then we define solution string of $a$ as

  "**Define** [name of a] **as** $var_0$; $var_1 = var_b + var_c = \cdots$ ; $var_2 = var_1 + var_d = \cdots$;
  **so** $var_0 = 7 \times var_2 = \cdots$."

  Here, the arithmetic computation is decomposed into 2-ary operations step by step separated with semicolons (so $\mathsf{op}_{G_\mathsf{d}}(a)$ is exactly the number of semicolons). The $var_0, var_1, var_2$ are three new (but distinct) random variables and their names are between a-z or A-Z and have 52 possible random choices. The "$\cdots$" ignores the math calculations.

- Given an abstract parameter $a \in$ Topo, suppose for instance $a = b \times c + d \times e + f \times g$ then we similarly define its solution text as

  "**Define** [name of a] **as** $var_0$; $var_1 = var_b \times var_c = \cdots$ ; $var_2 = var_d \times var_e = \cdots$;
  "$var_3 = var_f \times var_g = \cdots$ ; $var_4 = var_1 + var_2 = \cdots$; **so** $var_0 = var_3 + var_4 = \cdots$."

  Above, once again $var_0, var_1, var_2, var_3, var_4$ are new (but distinct) random variable names from a-z or A-Z, and we break down the computation into 2-ary operations.

With the above examples in mind, and combining those with real examples in Figure 11, it should be very clear how the solution texts are generated.

*Remark* D.6. $\mathsf{op}(G_\mathsf{d}^{\mathsf{nece}})$ is equal to the total number of semicolons in the solution text, because it represents the total (and minimum!) number of arithmetic operations needed to compute the final query parameter.

## D.4 Putting Altogether

We put together our data generation process for the structure graph $G_\mathsf{s}$ and the dependency graph $G_\mathsf{d}$ (along with $G_\mathsf{d}^{\mathsf{nece}}, \mathtt{Topo}$) in Algorithm 7.

In particular, we use global parameters $\mathsf{ip}_{\max}$ and $\mathsf{op}_{\max}$: the former controls the maximum number of instance parameters, and the latter controls the maximum number of solution operations. We select $n, m, s$ based on $\mathsf{op}_{\max}$ (to ensure that $1 \leq n \leq m \leq s \leq \mathsf{op}_{\max}$), and $d, e, w_0, w_1$ based on $\mathsf{ip}_{\max}$ and $s$. We also provide a boolean switch $force$ and when $\mathsf{force} = \mathsf{true}$, we shall force $s = \mathsf{op}_{\max}$ so that the generated math problem will have its solution to be of exactly $\mathsf{op}_{\max}$ operations.

We define datasets

- $\mathsf{iGSM}^{\mathsf{op} \leq \mathsf{op}_{\max}, \mathsf{ip} \leq \mathsf{ip}_{\max}}$ as the process of invoking $\mathtt{DrawAll}(\mathsf{op}_{\max}, \mathsf{ip}_{\max}, \mathsf{force} = \mathsf{false})$.
- $\mathsf{iGSM}^{\mathsf{op} = \mathsf{op}_{\max}, \mathsf{ip} \leq \mathsf{ip}_{\max}}$ as the process of invoking $\mathtt{DrawAll}(\mathsf{op}_{\max}, \mathsf{ip}_{\max}, \mathsf{force} = \mathsf{true})$.

Using this language:

- The training data iGSM-med is $\mathsf{iGSM}^{\mathsf{op} \leq 15, \mathsf{ip} \leq 20}$;
- The eval data of iGSM-med additionally includes $\mathsf{iGSM}^{\mathsf{op} = op, \mathsf{ip} \leq 20}$ for $op \in \{15, 20, 21, 22, 23\}$;
- The training data iGSM-hard is $\mathsf{iGSM}^{\mathsf{op} \leq 21, \mathsf{ip} \leq 28}$;
- The eval data of iGSM-hard additionally includes $\mathsf{iGSM}^{\mathsf{op} = op, \mathsf{ip} \leq 28}$ for $op \in \{21, 28, 29, 30, 31, 32\}$.

*Remark* D.7. During training (regardless of pretrain or finetune for probing tasks), we only use those data whose hash value of their solution template (see Footnote 11) is $< 17 \pmod{23}$, and during evaluation we only use those whose hash value is $\geq 17 \pmod{23}$. This ensures a strict separation between train and test data (even in terms of their solution templates).

*Remark* D.8. In Algorithm 7, we chose $s = \min\{t_0, t_1\}$, where $t_0$ and $t_1$ are two random integers between 1 and $\mathsf{op}_{\max}$. This choice encourages more easier math problems in the pretrain data, which we found improves the model's learning.

**Algorithm 7** `DrawAll(op_max, ip_max, force)` generation

1: $s \leftarrow \min\{t_0, t_1\}$ for $t_0, t_1$ being two random integers from 1 and $\mathsf{op}_{\max}$
2: If force = true then $s \leftarrow \mathsf{op}_{\max}$.
3: $n \leftarrow \max\{t_0, t_1\}$ for $t_0, t_1$ being two random integers from 1 and $s$
4: $m \leftarrow$ random integer between $n$ and $s$
5: $d \leftarrow$ a random choice among $\{2, 3, 4\}$ with distribution according to softmax(`weight`)
   ⋄ *for* `weight` $= [-(rel-0.2)^2, -(rel-0.5)^2, -(rel-0.8)^2]$ *for* $rel = \frac{s-1}{\mathsf{ip}_{\max}-1}$
6: $t_0, t_1 \leftarrow$ two random choices among $\{2, 3, 4\}$ with distribution according to softmax(`weight`)
7: $w_0 \leftarrow \min\{t_0, t_1\}$ and $w_1 \leftarrow \max\{t_0, t_1\}$.
8: $e \leftarrow \min\{t_0, t_1, (d-1)w_1^2\}$ for $t_0, t_1$ being random integers between $(d-1)w_0$ and $\mathsf{ip}_{\max}$
9: $G_{\mathsf{s}} \leftarrow \mathtt{DrawStructure}(e, d, w_0, w_1)$
10: $G_{\mathsf{d}}^{\mathsf{nece2}} \leftarrow \mathtt{DrawNecessary1}(G_{\mathsf{s}}, n, m)$
11: $(G_{\mathsf{d}}^{\mathsf{nece3}}, \mathtt{Topo}) \leftarrow \mathtt{DrawNecessary2}(G_{\mathsf{d}}^{\mathsf{nece2}})$ ⋄ *if fail, go to Line 9; if fail for 1000 times, go to Line 1*
12: $G_{\mathsf{d}}^{\mathsf{nece}} \leftarrow \mathtt{DrawNecessary3}(G_{\mathsf{d}}^{\mathsf{nece3}}, \mathtt{Topo}, s)$ ⋄ *if fail, go to Line 1*
13: $G_{\mathsf{d}} \leftarrow \mathtt{DrawUnnecessary}(G_{\mathsf{s}}, G_{\mathsf{d}}^{\mathsf{nece}})$
14: **return** $(G_{\mathsf{d}}, G_{\mathsf{d}}^{\mathsf{nece}}, \mathtt{Topo})$ ⋄ *and generate English descriptions following Section D.3*

## E   Data Details: Probing Data Preparation

We describe here how we prepare the probing data. We generate math data according to Appendix D.

For each problem and each probing task (such as `nece(A)`, `dep(A, B)`, etc), we need to specify two things: at which position to probe and what parameters $A$ (or $A, B$) to probe.

- For `nece` and `dep`, the probing always takes place at the end of the problem (and question) description, so there is no choice to be made; for `value`, `can_next`, `nece_next` tasks, the probing can take place at the end of *each* sentence in the solution for (including the beginning of the first solution sentence), and we uniformly at random make such choices.

- Each parameter $A$ (or $B$) can be uniformly at random chosen from the set of all (instance or abstract) parameters in our dependency graph $G_{\mathsf{d}}$ (with the only requirement that $A \neq B$).

In the end, we make sure for each problem and each probing task, we make at most 10 such random choices (over the position and the choice of parameters) and sample without replacement.

Just like in the pretrain data, we prepare our probing data so that only problems with hash values of their solution template (see Footnote 11) where the hash $< 17 \pmod{23}$ are included in the training set, and the rest are used for testing.

## F   Experiment Details

**Model.**   We use the GPT2 architecture [18], replacing its absolute positional embedding with modern rotary positional embedding [7, 20], still referred to as GPT2 for short. (We also played with the Llama architecture (especially with gated MLP layers) aand did not see any benefit of using it. This GPT2 performs comparably to Llama/Mistral at least for knowledge tasks [4].)

Let GPT2-$\ell$-$h$ denote an $\ell$-layer, $h$-head, $64h$-dim GPT2 model. We primarily use GPT2-12-12 (a.k.a. GPT2-small) in this paper, but in Section 6 we explore larger models with different widths and depths. Our size-1 models are GPT2-4-21, GPT2-8-15, GPT2-12-12, GPT2-16-10, GPT2-20-9, roughly the same size as GPT2-small. Our size-2 models are GPT2-4-30, GPT2-8-21, GPT2-12-17, GPT2-16-15, GPT2-20-13, roughly twice the size of GPT2-small. We use a context length of 768/1024 for language model pretraining on iGSM-med/iGSM-hard and a context length of 2048 for evaluation.

**Data size.** For both pretraining and finetuning, we did not limit the amount of training data; we generated new data on-the-fly. We do not explore sample complexity in this paper, such as the number of math problems needed to achieve a certain level of accuracy, as it would complicate the main message of this paper.

## F.1 Pretrain Experiment Details

**Pretrain parameters.** We used the AdamW optimizer with mixed-precision fp16, $\beta = (0.9, 0.98)$, cosine learning rate decay (down to 0.01x of peak learning rate in the end), and 1000 steps of linear ramp-up. We used a mixture of V100/A100 GPUs, but the GPU specifications are not relevant here.[27] For all of our pretrain experiments:

- On the iGSM-med datasets, we used a (peak) learning rate 0.002, weight decay of 0.05, batch size of 512, context length of 768, and trained for $100,000$ steps.

- On the iGSM-hard datasets, we used a (peak) learning rate 0.002, weight decay of 0.03, batch size of 256, context length of 1024, and trained for $200,000$ steps.

Our pretrain data is constructed by randomly generating math problems (and solutions), concatenating them together, and truncating them (in the right) to fit within the 768 or 1024-sized context window. If a problem is longer than the context window size, we discard it (this happens very rarely).

**Test-time parameters.** When evaluating on test data, we use context length 2048 for both iGSM-med and iGSM-hard. We use either beam=1 and dosample=False (greedy) or beam=4 and dosample=True (beam-search multinomial sampling) to present test accuracies. We discover it is better to keep dosample=False while beam=1 and dosample=True while beam=4. We also tried larger beam sizes and found no further improvements.

**Accuracy statistics.** Our main accuracies are presented in Figure 3, where each entry is averaged over 4096 math problems of that type. Our accuracies *are not simply* from comparing the answer integers (between 0 and 22); instead we have written a parser to make sure the model's intermediate solution steps are fully-correct.

For the "redundancy" experiment Figure 4, we tested each model again with 4096 math problems in each case and presented the results among fully-correct solutions. For this figure, we present beam=1 for cleanness and the results for beam=4 are almost completely identical.

For the "depth matters" experiment Figure 9, because we care about the (relatively small) accuracy differences across models, we pretrain using two different random seeds, and evaluate with both beam=1/4; we then present the best accuracies in each entry with respect to the 2 seeds and 2 beam choices. The accuracies are again over 4096 math problems.

## F.2 V-probing

Our V-probing was first introduced in Section 4.1 with more details given in Section B. It is a *fine-tuning* process upon the pretrained language model, with an additional linear head on the output layer, and a small rank-$r$ update on the input (embedding) layer. The pretrained model is freezed, and only this linear head and the rank-$r$ update are trainable parameters during the fine-tuning.

---
[27]A 128-GPU job with batch size 1 each would be identical to a 32-GPU job with batch size 4 each.

Recall we use $r = 8$ in this paper (in contrast, the hidden dimension of GPT-12-12 is 768). This small value of $r$ ensures if probing accuracy is high, it mostly comes from the pretrained model and not the additional trainable parameters.

For V-probing, we use the same configurations as pretrain, except that:

- For V-probing on the iGSM-med datasets, we used a learning rate of 0.002 (with no ramp-up, linear decay down to 0), weight decay of 0.01, batch size of 256, and trained for 100,000 steps.

- For V-probing on the iGSM-hard datasets, we used a learning rate of 0.002 (with no ramp-up, linear decay down to 0), weight decay of 0.01, batch size of 128, and trained for 100,000 steps.

**V-probing statistics.** In Figure 7(a), Figure 7(b), Figure 12, Figure 8(a), and Figure 8(b), we tested at least 4096 random problem-parameter pairs *in each cell*. In Figure 8(a) and Figure 8(b), when evaluating probing results on GPT-2 model's *generated* correct or wrong solutions, we used beam=1 and dosample=False (greedy) for generation. (Results are similar for beam=4.)

In our layer-wise nece($A$) probing experiments (Figure 10 and Figure 14), we tested at least 73728 random problem-parameter pairs in each case and then divided the results into bins based on the parameter $A$'s distances to the queries.

## G  Failure Examples on GPT-4 / GPT-4o

In Figure 2, we conduct few-shot experiments using the latest versions of GPT-4 turbo (2024-04-09) and GPT-4o (2024-05-13) models to evaluate their accuracies on our iGSM-med$_{pq}$ dataset, with respect to different op $\in \{2, 3, \ldots, 20\}$.

To ensure meaningful evaluation:

- We replaced mod23 with mod5 to ensure that any errors are not due to arithmetic mistakes. We also provided a few arithmetic computation examples.

- We minimized English diversity to ensure that any errors are not due to misunderstanding the problem description. Specifically,

  - We fixed a simple categorization (School, Classroom, Backpack, Stationerys), with only four items in each category.
  - We provided an English *background* paragraph to fully describe the structure graph (i.e., which item has which subitem), as well as the number of items in each category. The math problem is preceded by this background paragraph.

- We provided five-shot problem/solution examples to ensure that GPT-4 understands how to solve such math problems step by step.

We did not verify each step of GPT-4's solution but checked if the final output number (between 0 and 4) matched the correct answer. The accuracy results are presented in Figure 2. It shows that the GPT-4o model is almost randomly guessing for op $\geq 11$, and GPT-4 turbo for op $\geq 9$.

Furthermore, Figure 15 shows that when the GPT-4/4o models fail to answer the math problems, it is mostly not due to format errors or misunderstanding of the problem. Instead, just like what we discovered in Section 5, GPT-4/4o fail also because they compute unnecessary parameters (i.e., nece($A$) = false) or compute parameters that are not yet ready to be computed (i.e., can_next($A$) = false). This further confirms that our findings do connect to practice, regarding the model's hidden reasoning process.

Figure 15: Failure examples for GPT-4/GPT-4o on iGSM-med$_{pq}$. They make mistakes similar to what we discover in this paper, that is to compute unnecessary parameters in the solutions (i.e., $\texttt{nece}(A) = \text{false}$), as well as computing parameters that are not yet ready to compute (i.e., $\texttt{can\_next}(A) = \text{false}$).